\documentclass{article}

\usepackage[nonatbib, final]{neurips_2019}

\usepackage[utf8]{inputenc} 
\usepackage[T1]{fontenc}    
\usepackage{hyperref}       
\usepackage{url}            
\usepackage{booktabs}       
\usepackage{amsfonts}       
\usepackage{nicefrac}       
\usepackage{microtype}      
\usepackage[table,xcdraw]{xcolor}
\usepackage{todonotes}
\usepackage{subfiles}
\usepackage{verbatim}
\usepackage{subcaption}
\mathchardef\mhyphen="2D
\title{AI and Wargaming}

\author{
  James Goodman \\
  Queen Mary University of London\\
  \And
    Sebastian Risi \\
  IT University of Copenhagen\\
     \And
  Simon Lucas \\
  Queen Mary University of London\\
}

\begin{document}

\maketitle

\tableofcontents

\section{Executive Summary}
Recent progress in Game AI has demonstrated that 
given enough data from human gameplay, or experience gained 
via simulations, machines can rival or surpass even the most skilled human players in some of the most complex and tightly contested games.
The question arises of how ready this AI is to be applied to wargames.

This report provides a thorough answer to that question, summarised as follows.

\begin{itemize}
    \item Wargames come in a number of forms --- to answer the question we first clarify which types we consider.
    \item In order to relate types of wargames to the performance of AI agents on a number of well known games, such as Go and StarCraft, we provide the most comprehensive categorisation to date of the features of games that affect the difficulty for an AI (or human) player.
    \item In the last few years some amazing results have been demonstrated using Deep RL (and Monte Carlo Tree Search) on games such as Go, StarCraft and Dota 2.  We review the main architectures and training algorithms used, the level of effort involved (both engineering and computational) and highlight those which are most likely to transfer to wargames.
    \item All the most impressive results require the AI to learn from a large number of game simulations.  Access to a fast and copyable game engine/simulator also enables statistical forward planning algorithms such as Monte Carlo Tree Search and Rolling Horizon Evolution to be applied.  These should be considered as they provide intelligent behaviour ``out of the box'' i.e. with no training needed, and can be combined with learning methods such as Deep RL to provide even more intelligent play.
    \item Explainable decision making can be achieved to some extent via the visualisation of simulations, and by analysing neural network activation patterns to help explain the operation of Deep RL systems.  Explainability is best seen as desirable rather than essential.
    \item There is a strong need for a software framework tailored towards wargame AI.  There are many examples of successful game AI frameworks, and how they can provide a significant boost to a research area.  Whilst no existing one provides adequate support for wargames, we make clear recommendations on what is needed.
\end{itemize}

\vfill
\pagebreak

\section{Introduction}\label{sect:Intro}

This study was commissioned by the Defence Science and Technology Laboratory (Dstl), part of the UK Ministry of Defence. In light of major advances in AI techniques in recent years, notably the success of Deep Reinforcement Learning (Deep RL) in achieving championship level play in such games as Go and Starcraft II, Dstl and its international partners would like to apply artificial intelligence techniques to military wargames to allow them to deliver outputs more quickly, cheaply and/or of higher quality.

Wargames are used to inform decisions on future force structures, future military capabilities (equipment, together with supporting lines of development), and to prepare for operations.
Wargames can be conducted in lots of different ways, ranging from seminar wargames, through manual board games to complex computer-assisted wargames, where the computer adjudicates the outcomes of engagements. 

Specific goals include using AI to discover tactics that could improve operational outcomes using existing military capabilities, or could suggest effective concepts of use for new military capabilities under consideration.

The headline successes of Deep RL have often required large computational and other costs; rough estimates are \$250,000+ for a single training run of the AlphaZero Chess/Go/Shogi agent, and several million dollars for the Starcraft II results, once the cost of the development team is taken into account. 
Through this lens it is also vital to focus limited resources on areas of AI in wargames where they can achieve results without mandating the same level of investment for each individual game.

This report reviews the current state of the art of AI and Machine Learning applied to games and similar environments to identify:
\begin{enumerate}
    \item {What techniques from recent academic advances are applicable to wargames? The academic literature has focused on abstract games, such as Chess or Go, or commercial games, such as Space Invaders, Dota2 or Starcraft II. These are not games that are \textit{directly} relevant to real wargames.}
    \item{What problems in actual wargames are tractable by current AI techniques?}
\end{enumerate}

We first consider the similarities and differences between wargames and some of the environments in which recent AI research has been conducted (Section~\ref{sect:GameFeatures}). We then review recent progress in Deep RL and other techniques in Sections~\ref{sect:DeepRL} and ~\ref{sect:OtherAlgos} before drawing out in Section~\ref{sect:algoSummary}  aspects and techniques from recent Game AI and related research most relevant to wargames given the analysis in Section~\ref{sect:GameFeatures}. 
Section~\ref{sect:Recommendations} concludes with Recommendations for next steps to apply AI to wargames.

\subsection{Types of wargames}\label{sect:wargameTypes}
Wargames are used by defence and military organizations for several distinct purposes. 
\begin{enumerate}
    \item Planned Force Testing. These are large-scale wargames involving tens to hundreds of participants in multiple different and collaborative commands. They are concerned with end-to-end testing of holistic military capabilities across logistics, diplomacy, theatre access and scenario aftermath as well as combat. Computer tools may be used for some of the detailed moderation, but these are primarily run as analogue exercises.
    \item Plan Variation Testing. This takes a single scenario plan, for example part of an outcome from Planned Force Testing, to consider in more detail. These are often run multiple times to analyse interesting branch-points that need human decision-making focus.
    \item Concept/Force Development. These wargames take a proposed concept or change to capability, such as a new rifle or autonomous land vehicle, and explore the associated strategy/tactics space to find viable concept combinations and potential doctrine suitable for the proposal. They will be at the minimum scale needed to test the proposal.
    \item Procurement. These wargames are the most constrained type. They are used in the context of a decision to purchase one type of equipment from a small set of options. The scenarios are tightly controlled and repeated multiple times to quantify as far as possible the differences between the options. Ideally these outcome differences would be analysed with changes to equipment but not military doctrine; with doctrine changed by old equipment; and with both doctrine and equipment changes.
\end{enumerate}

This study focuses on the potential use of AI techniques in computer-moderated wargames. 
These tend to be complex games with large state spaces and large action spaces, to name just two dimensions of complexity.  Section~\ref{sect:GameFeatures}  provides a comprehensive study of all the main aspects of a game that can make it difficult for an AI or human player to play it well.  This focus on complex games is in contrast to abstract game-theory type games where the actions and rewards for each player form a payoff matrix, and the games
have no intrinsic state.



\vfill
\pagebreak

\section{The Features of Wargames} \label{sect:GameFeatures}
In this section we review the core features of wargames, comparing and contrasting them with those of game environments commonly used in AI and Machine Learning research.
We focus specifically on the potential use of AI techniques in \emph{computer-moderated} wargames, with some comments on techniques relevant to support physically moderated wargames made where appropriate.


We consider some concrete games of interest:
\begin{itemize}
    \item {Command Modern Air and Naval Operations (CMANO)\footnote{https://www.matrixgames.com/game/command-modern-operations}}
    \item {Flashpoint Campaigns\footnote{https://www.matrixgames.com/game/flashpoint-campaigns-red-storm-players-edition}}
    \item {ODIN, CAEn}
\end{itemize}
The first two of these are commercial wargames in the air/sea and land domains respectively, but with sufficiently accurate simulation detail to be useful for as a moderation tool to determine possible detailed combat outcomes in a larger wargame, concept filtering, or for quick branch analysis in a larger game to determine if this is `interesting' in some way.
The third set are computer-moderated wargame environments for the defence community; CAEn being developed in-house by Dstl, and ODIN a commercial defence package from Atlas Elektronik.

As a comparison we consider the following games that are used frequently in recent AI research:
\begin{itemize}
    \item {Go/Chess. These represent classic 2-player turn-based abstract games.}
    \item {Starcraft II. A complex commercial Real-Time Strategy (RTS) game.}
    \item {Atari/GVGAI. A diverse set of relatively simple games inspired by early commercial computer games of the 1980s that was the basis of the first Deep RL breakthroughs in 2015 by DeepMind~\cite{atari}.}
    \item {MicroRTS ($\mu$RTS). A simplified RTS used for academic research~\cite{microRTSAlphaGo}. MicroRTS encapsulates the core of a RTS game (real-time control of multiple units, with resource gathering and a build/tech tree), with an interface designed for AI research and a rudimentary GUI.}
    \item{Dota 2. A commercial team-based game with five separate players on each team controlling a hero character. Each team seeks to destroy the opponent team's base to win.}
    \item {MuJoCo. A physics simulation environment, in which a policy learns to control robots of varying complexity. This is not a game, but is a popular environment for recent Deep Learning research as well as comparisons with Evolutionary Algorithms~\cite{Bansal_Pachocki_Sidor_Sutskever_Mordatch_2017, Salimans_Ho_Chen_Sidor_Sutskever_2017, Shyam_Jaskowski_Gomez_2019, Tong_Liu_Li_2019}}
\end{itemize}

\begin{table}[]
\begin{tabular}{|l|l|l|l|l|l|l|l|}
\hline
\textbf{Feature}  & \textbf{Ref}      & \textbf{Chess}  & \textbf{Atari/} & \textbf{Starcraft}    & \textbf{$\mu$RTS} & \textbf{Dota2} & \textbf{MuJoCo} \\ 
    &  & \textbf{/Go}  & \textbf{GVGAI} &   &  &  & \\ \hline
Action Space & \ref{sect:ActionSpace}   & \cellcolor[HTML]{FE0000}  &  \cellcolor[HTML]{FE0000}   & \cellcolor[HTML]{34FF34} &  \cellcolor[HTML]{F8A102}&  \cellcolor[HTML]{F8A102} &  \cellcolor[HTML]{FE0000}    \\ \hline
Branching Factor  & \ref{sect:Branching}      & \cellcolor[HTML]{FE0000}   & \cellcolor[HTML]{FE0000}  & \cellcolor[HTML]{34FF34} &  \cellcolor[HTML]{34FF34}& \cellcolor[HTML]{34FF34} & \cellcolor[HTML]{FE0000}   \\ \hline
Number of Decisions  & \ref{sect:NumberDecisions}   & \cellcolor[HTML]{F8A102}   & \cellcolor[HTML]{F8A102} &\cellcolor[HTML]{F8A102} & \cellcolor[HTML]{F8A102}& \cellcolor[HTML]{F8A102}  & \cellcolor[HTML]{F8A102}  \\ \hline
State Space    & \ref{sect:StateSpace}         & \cellcolor[HTML]{FE0000} & \cellcolor[HTML]{FE0000}  & \cellcolor[HTML]{34FF34} & \cellcolor[HTML]{34FF34}& \cellcolor[HTML]{34FF34}  &  \cellcolor[HTML]{F8A102} \\ \hline
Observability   & \ref{sect:Observability}        & \cellcolor[HTML]{FE0000}& \cellcolor[HTML]{FE0000}  & \cellcolor[HTML]{34FF34} & \cellcolor[HTML]{FE0000}& \cellcolor[HTML]{34FF34}  & \cellcolor[HTML]{FE0000}  \\ \hline
Observation Space  & \ref{sect:ObservationSpace}     & \cellcolor[HTML]{FE0000} &\cellcolor[HTML]{FE0000}  & \cellcolor[HTML]{F8A102} & \cellcolor[HTML]{F8A102}& \cellcolor[HTML]{F8A102} & \cellcolor[HTML]{FE0000} \\ \hline
Information Flow   & \ref{sect:InfoFlow}        & \cellcolor[HTML]{FE0000} & \cellcolor[HTML]{FE0000}  & \cellcolor[HTML]{FE0000} & \cellcolor[HTML]{FE0000}& \cellcolor[HTML]{FE0000}  & \cellcolor[HTML]{FE0000}\\ \hline
Stochasticity  & \ref{sect:Stochasticity}         & \cellcolor[HTML]{FE0000} & \cellcolor[HTML]{FE0000}  & \cellcolor[HTML]{34FF34} & \cellcolor[HTML]{FE0000}& \cellcolor[HTML]{34FF34} & \cellcolor[HTML]{FE0000}\\ \hline
Win conditions   & \ref{sect:WinConditions}       & \cellcolor[HTML]{FE0000}  &  \cellcolor[HTML]{34FF34} & \cellcolor[HTML]{F8A102} &  \cellcolor[HTML]{F8A102}& \cellcolor[HTML]{F8A102} & \cellcolor[HTML]{F8A102} \\ \hline
Reward Sparsity & \ref{sect:RewardSparsity}          & \cellcolor[HTML]{FE0000}  &  \cellcolor[HTML]{34FF34} & \cellcolor[HTML]{34FF34} &  \cellcolor[HTML]{34FF34}& \cellcolor[HTML]{F8A102} & \cellcolor[HTML]{F8A102} \\ \hline
Opponents   & \ref{sect:Opponent}       & \cellcolor[HTML]{34FF34} &  \cellcolor[HTML]{FE0000} & \cellcolor[HTML]{34FF34} &  \cellcolor[HTML]{34FF34} & \cellcolor[HTML]{34FF34} & \cellcolor[HTML]{FE0000} \\ \hline
Scenario Variability & \ref{sect:ScenarioVar} & \cellcolor[HTML]{FE0000}  & \cellcolor[HTML]{34FF34}  & \cellcolor[HTML]{F8A102} & \cellcolor[HTML]{34FF34}& \cellcolor[HTML]{F8A102}  &  \cellcolor[HTML]{FE0000} \\ \hline
Objective Variability & \ref{sect:ObjectiveVar}  & \cellcolor[HTML]{FE0000}  & \cellcolor[HTML]{FE0000} & \cellcolor[HTML]{FE0000} & \cellcolor[HTML]{FE0000}& \cellcolor[HTML]{FE0000} & \cellcolor[HTML]{FE0000}  \\ \hline
Player Asymmetry  & \ref{sect:PlayerAsymmetry}      & \cellcolor[HTML]{FE0000}  & \cellcolor[HTML]{FE0000} & \cellcolor[HTML]{34FF34} & \cellcolor[HTML]{F8A102} & \cellcolor[HTML]{34FF34}  & \cellcolor[HTML]{FE0000} \\ \hline
\end{tabular}
\caption{Summary of similarities of common AI research environments to wargames. Green-Amber-Red indicates a rough gradation from Similar to Dissimilar.}
\label{SummaryTable}
\end{table}

Table~\ref{SummaryTable} summarises the similarity of these games to wargames. 
It is immediately visible from Table~\ref{SummaryTable} that Dota 2, Starcraft and RTS games are more similar than the other options to wargames. MuJoCo is a robotic control environment and not surprisingly is the least similar. The reason for its inclusion at all is that it is a very common environment for Deep RL research, especially in the policy optimization literature. The point to make is that work in this domain should be used with caution in wargames (but note that the PPO algorithm used by OpenAI in their Dota 2 work was originally used in MuJuCo, so this work is still useful with appropriate modifications).
The Atari/GVGAI games are less similar to the wargame domain in aggregate, but in some areas (Win Conditions and Scenario Variability) are better models for wargames than either Dota 2 or Starcraft.

In order to provide a systematic analysis
we develop an extensive set of features\footnote{This is partly inspired by
Murray Campbell's IEEE CIG 2017 keynote
talk section on ``Characterizing Games (work in progress)'', and we thank Murray for sharing his slides.}
on which to compare each game.  While
this is intended to be a comprehensive 
and ultimately an exhaustive set of features,
it is currently a work in progress.
Although
we have developed the framework with Wargames in mind, it is readily applicable to other types
of game.

The detail behind each of the component features is expanded in the following sections. 
Each section briefly discusses specific AI techniques that address the corresponding feature and these comments are then pulled together in Section~\ref{sect:algoSummary} to highlight the algorithms and areas of the academic literature most relevant to wargames.  

Since proposing and applying the current
set of features we have recently identified
more that could be used, but have not applied
them in this report:

\begin{itemize}
\item Rule complexity (e.g. as measured by some minimum description length measure of the game rules).
\item Human knowledge of game (this can affect how the quality of AI play will be judged).
\item Availability of game-play data (this is useful for imitation learning).
\end{itemize}

Note that a key aspect of game difficulty for
AI is whether we have access to a fast and copyable
forward model.  We have not included it in the
set of features below because it is not strictly speaking part of the game.

\subsection{Action Space}\label{sect:ActionSpace}
The Action Space of a game is the set of actions available to a player when they make a move. In the case of Go this is a discrete set of 361 board locations (on a 19 by 19 board), and in Atari games this is the set of possible joystick positions, plus the pressing of the button (a maximum of 18 for all combinations, but most game use far fewer). These number represent an upper bound as not all of these actions will be valid in any given situation, for example in Go some of the board positions will already be occupied and hence unavailable.

A key distinction is whether the action space of a game is \textbf{discrete}, as in the Go and Atari examples, or \textbf{continuous}. In MuJoCo for example an `action' is the setting of angles and torques for all limb rotations and movements, and each angle can be any value between 0 and 360 degrees. In this the number of dimensions that need to be specified is important. In games such as Starcraft II or CMANO, any specific unit can be ordered to move to a map position, which is a continuous two or three dimensional vector; although other actions, such as the target of an attack, are discrete.

Any continuous action space can be turned into a discrete one by selecting specific points as chooseable (`discretisation'). This is part of the design of MicroRTS with the game working on a fixed grid, hence discretizing a continuous 2-dimensional space into fixed squares; Flashpoint, like other wargames, does the same with hexes.

Table~\ref{table:ActionSpace} summarises the features of the different environments. Starcraft II and MicroRTS are similar in style, with multiple units being controlled simultaneously and hence a full action space that is combinatorially very large. Chess, Go, Atari and GVGAI are a different type of environment with a smaller set of actions available at any one time. A player in Dota 2 is halfway between these, and controls a single unit which can take one action at a time (but up to 80,000 actions may be possible).

The Starcraft II action space can be considered in two ways:
\begin{enumerate}
    \item {Clicks on screen. This is the approach used by DeepMind with Deep Learning, and the gives the rough $10^8$ set of actions each turn, as these are defined by the human point-and-click mouse interface~\cite{vinyals2017starcraft}.}
    \item{Unit orders. This uses an API, which provides a list of available units and the orders that can be given to each.}
\end{enumerate}
The second of these is most appropriate for wargames, for which there is no requirement to additionally learn the vagaries of a human interface.

Classic AI approaches, including MinMax and Monte Carlo Tree Search as well as Q-Learning based approaches used in Go and Atari require a discretised action space. Policy Gradient RL techniques have been developed to operate with a continuous action space (or policy)~\cite{policyGradient}. Another approach is to discretise the space, and this is also used to reduce the sizes of discrete spaces to a smaller number. This `Action Abstraction' reduces a large number of possible actions to a small set tractable for forward planning or other techniques~\cite{Churchill_Buro_2013, Moraes_Marino_Lelis_Nascimento_2018}.
For example in some MicroRTS and Starcraft II work, a set of scripts can be used to define a particular tactic (`worker rush', `build economy', 'defend against incoming attack'), and the AI focuses on learning which tactic to apply in a given situation without needing to learn the detailed instructions that make it up~\cite{Churchill_Buro_2013, Neufeld_Mostaghim_Perez-Liebana_2019}.
Sub-goal MCTS does something similar with pre-specified sub-goals that are used to automatically prune the action space~\cite{Gabor_Peter_Phan_Meyer_Linnhoff-Popien_2019}.

This is a key mechanism to introduce domain knowledge in wargames. It can leverage expert knowledge embedded in the scripts that define useful action sequences (such as `take cover', `seek high ground', `patrol with active sonar'), without being fully constrained as the AI can learn which script to use in which situation, which may not always accord with expert intuition. The net result is to significantly speed up the learning or planning process (by orders of magnitude) at a cost in loss of flexibility.

Extensions of this idea relevant to wargames are to simultaneously learn the parameters of scripts (for example, an aggression parameter that specifies the odds required to attack); to evolve a sub-set of scripts to use from a larger population~\cite{Marino_Moraes_Toledo_Lelis_2019}; or to use the results of scripted recommendations for more detailed low-level planning~\cite{Barriga_Stanescu_Buro_2017}. This last work generates actions for units based on scripts, and then devotes some of the computational budget to refining proposed moves for units that are close to enemy units, for which a small modifications of a high-level default move may have an especially large impact. 
(There is overlap here with the unit-based techniques discussed in Section~\ref{sect:unitTechniques}.)

\begin{table}[]
\begin{tabular}[t]{|l|l|c|c|c|}
\hline 
\textbf{Game} & \textbf{Discrete/Cont.} & \textbf{Order Mode}  & \textbf{Max Action Space} & \textbf{Decisions} \\
\hline
\textbf{CMANO} & Continuous & $10^{0\mhyphen 3}$ units & $10^{10+}$ & $10^5$ \\
\textbf{Flashpoint} & Discrete & $10^{1\mhyphen 2}$ units &  $10^{10+}$ &$10^0$ \\
\textbf{Chess/Go} & Discrete & 1 move &  $10^2$  to $10^3$ &$10^0$ to $10^1$ \\
\textbf{Atari/GVGAI} & Discrete & 1 move &   $10^1$ & $10^2$ to $10^4$ \\
\textbf{Starcraft II} & Continuous & $10^{0\mhyphen 3}$ units & $10^8$  or $10^{10+}$ & $10^{5+}$ \\
\textbf{Dota2} & Discrete & 1 move & $10^5$  & $10^{5+}$ \\
\textbf{MicroRTS} & Discrete & $10^{1\mhyphen 2}$ units &  $10^6$ to $10^8$ & $10^2$ to $10^3$ \\
\textbf{MuJoCo} & Continuous & 3-17 dimensions &   - & $10^{3+}$  \\
\hline 
\end{tabular}
\caption{Action Space Categorisation. `Order Mode' can be 1 move per turn, orders per unit or a single multi-dimensional vector per time-step. `Decisions' is an approximation of the number of decisions a player makes during one game.}
\label{table:ActionSpace}
\end{table}
\subsection{Branching Factor}\label{sect:Branching}
The Branching Factor of a game is the number of new positions that can be reached after each move/action. In general the higher this is, the more challenging the game is to any AI technique.
This is closely related to the Action Space. In a deterministic game, each distinct action will lead to a different game position and hence the branching factor is equal to the average action space size. In a stochastic game then the branching factor can be much higher, for example in a wargame a decision to attack a unit can lead to many different states depending on the random damage dealt to both sides, and the results of any morale checks.
The impact of branching factor is hence covered in Sections~\ref{sect:ActionSpace} and ~\ref{sect:Stochasticity} on Action Space and Stochasticity respectively.

\subsection{Number of Decisions}\label{sect:NumberDecisions}

This refers to the average number of decisions or actions
made by a player during the game.  Games with more decisions are potentially more complex, but this is also dependent on the size of the action space.
For this reason the number of decisions a player makes is included alongside Action Space in the third line of Table~\ref{table:ActionSpace}, as these two can roughly be multiplied together as a rough gauge of complexity.

The length of a game can be characterised by either the number of decisions
made by each player or the number of game simulation
`ticks'.  Classic games such as chess have one `game tick' per decision.
Realtime wargames may play out for tens of thousands of game simulation ticks, 
but on most of those the players may be taking no actions (even if they could)
due to limits on human perception, thinking and reaction times.  In Starcraft,
the AI agent may interact via a constrained interface in order to reduce
the click rate and level the playing field when playing against human players.

The very different numbers of decisions available for CMANO ($10^5$) and Flashpoint ($10^0$ to $10^1$) in Table~\ref{table:ActionSpace} draw attention to two quite different ways of looking at wargames.
The figure for CMANO is estimated from a maximum of one action per second over a 2-3 hour realtime game; which is of the same order of magnitude as in Starcraft and Dota 2. 
In Flashpoint Campaigns the player has a `command cycle' that only allows them to enter orders (for all units) at a few specific points. Once orders have been given, 15-30 minutes of simulated time then unfold during which the player can only watch events.

This difference is between a low-level versus high-level control perspective, which can apply to \emph{either} an AI or human player. Flashpoint seeks to model the constraints around human decision making in the field, in line with much of the purpose of wargaming outlined in Section~\ref{sect:wargameTypes}. The second by second view is more in line with low-level control of units to fulfil these goals, and in Flashpoint these are in effect executed by the in-game AI that makes decisions for each unit about moving, attacking and retreating based on the high-level goals provided by the player. This perspective is relevant for AI that supports a strategic human player by micromanaging units for them in pursuit of a specified high-level goal.
Neither perspective is similar to the strict turn based approach of classic games, or the Atari/GVGAI environment with a very constrained set of actions at each decision point.

This distinction between low-level (or `tactical') and high-level (or `strategic') decisions is more important than the raw numerical differences in Table~\ref{table:ActionSpace}. It is addressed by a number of techniques covered in more detail elsewhere in this report, specifically:
\begin{itemize}
    \item Hierarchical methods in Sections~\ref{sect:unitTechniques} and~\ref{sect:HRL};
    \item Action Abstraction (for example each high-level strategy being a script that specifies the low-level tactics to follow) in Section~\ref{sect:ActionSpace}.
\end{itemize}

\subsection{State Space}\label{sect:StateSpace}
\begin{table}[]
\begin{tabular}[t]{|l|l|l|l|l|l|l|}
\hline 
\textbf{CMANO} & \textbf{Flashpoint} & \textbf{Chess/Go}  & \textbf{Atari/} & \textbf{Starcraft II}    & \textbf{MicroRTS} & \textbf{MuJoCo}\\ 
 &  &   &  \textbf{GVGAI} & \textbf{(+ Dota2)}  &  & \\ 
\hline 
LC & LC & $10^{47, 170}$ & $10^{6000}$ / LC & $10^{1685}$ / LC & LC & LC \\
Unit &  Unit & Global  & Global  &  Unit & Unit  & Global\\
\hline
\end{tabular}
\caption{State Space categorization. `LC' indicates a game is `locally continuous'. The state-space for Atari is based on the possible values for all pixels over four frames after downsampling~\cite{atari}. State-space estimates for other games from~\cite{Ontanon_2017}. The second line indicates if state is purely global, or if it can largely be factored to individual units.}
\label{table:StateSpace}
\end{table}
We can categorise the State Space of a game in terms of the number of different states it can be in. These are calculable in cases like Chess or Go ($10^{47}$ and $10^{170}$ respectively). In games with continuous state spaces, these are technically infinite.
This categorisation is especially relevant to search-based techniques, as used classically for Chess. 
In Chess or Go, each board position is unique (ignoring symmetries), and it can make a huge difference to the outcome of a game if a pawn is one square to the left, or if a Go stone is moved by one space. A difference of one square for a pawn can for example can open up an attack on a player's Queen or King.

This is much less true in Starcraft II, many Atari games, and RTS games in general. These games are more `locally continuous', in the sense that small changes in unit position, health or strength usually lead to similarly small changes in the outcome. 

We should also distinguish between the state space and the observation space (section~\ref{sect:ObservationSpace}).  The observation
space is the lens through which the agent views the underlying state.  The difference between state and observation of state is emphasized for partially observable games, where the agent is forced to make
assumptions about the unobservable parts of the state.
However, it may also be that the observation is an
expanded version of the underlying state.

For example the full observation space for Atari 2600
games is $10^{70,000}$ based on the number of pixels, the possible colours of each pixel. This is reduced (downsampled) to $10^{6000}$ for processing by the deep network in the original (and later) DeepMind work~\cite{atari}. This downsampling by thousands of orders of magnitude `loses' all but an infinitesimal fraction of the original information without noticeably reducing performance, and is a clear indication that the game is locally continuous in this sense.
In this regard they are wargame-like, and classic state-space complexity is less relevant.

However, the Atari 2600 console only has 128k bytes of RAM (1024 bits) so the underlying state space is limited
by this, and hence much smaller than even the 
compressed observation space described above (though some cartridges came with up to 4k
of extra RAM).  The main point is that after graphical
rendering the observation space can be much
larger than the underlying state space.

Local continuity still allows for discontinuities; for example an infantry company may move only a few metres, but if this is from inside a dense wood onto an open plain in sight range of the enemy then it will have a major, discontinuous impact on the game. The key point is that the formal state space size is not a good measure of game complexity. 
It is better to think about this as a low-dimensional, non-linear manifold within the high-dimensional pixel (or other) space. 
This lower-dimensional manifold represents the true complexity of the state space, but is not directly observable.
It is this low-dimensional manifold that machine learning techniques seek to identify, building a model that understands the boundary between wood and plain is the important implicit variable.
Deep RL does this using multiple convolutional neuron layers, random forests by constructing multiple decision trees and so on.

Hence in Table~\ref{table:StateSpace}, we categorize games by this idea of `local continuity'. In addition to a distinction between locally-continuous and discrete games, some games can be largely factored into the state of individual units. In these cases, we can consider the state space to be composed of a number of distinct components:
\begin{enumerate}
    \item {The state of each unit; position, damage, equipment etc.}
    \item {The interactions between units; their relative positions are likely to be relevant in a wargame.}
    \item {Purely global state; developed technologies in Starcraft II, or the number of available air-strikes in Flashpoint.}
\end{enumerate}
Wargames are unit-based and locally continuous, and most similar to the RTS-style games in Table~\ref{table:StateSpace}.
The number of units that a player controls is an important aspect of the complexity of RTS games and wargames.
A large number of units are in play limits the 
ability of a player to track them all, and to properly consider
the best actions that each should take in relation to what
the others are doing.  In wargames such as CMANO, a player can
even write Lua scripts to control how a set of units behave.

As the number of units increase, so the need for some type
of structured system of control becomes apparent, such
as standard military hierarchies.
Various approaches have been used in the academic literature looking at unit-oriented state spaces, and these are surveyed in detail in Section~\ref{sect:algoSummary} given how important this aspect is to wargames.

\subsection{Observability}\label{sect:Observability}
\begin{table}[]
\begin{tabular}[t]{|l|l|l|l|l|l|l|l|}
\hline 
\textbf{CMANO} & \textbf{Flashpoint} & \textbf{Chess}  & \textbf{Atari/} & \textbf{Starcraft II} & \textbf{Dota2}     & \textbf{$\mu$RTS} & \textbf{MuJoCo}\\ 
 &  &  \textbf{/Go} &  \textbf{GVGAI} & &    &  & \\ 
\hline 
Imperfect & Imperfect & Perfect & Perfect & Imperfect & Imperfect & Perfect & Perfect \\
Fog of War &  Fog of War &   &   &  Fog of War&  Fog of War &   & \\
Unit & Unit & & & Camera/Unit & Unit & &  \\ 
\hline
\end{tabular}
\caption{Observability categorization. The first line shows if a game has Perfect or Imperfect information, the second line the type of Imperfect information, and the third line how Observability is restricted.}
\label{table:Observability}
\end{table}
The Observability of game defines what elements of the full game state are visible. At one extreme, a game with Perfect Information has the whole state space visible to all players. 
A game with Imperfect Information has some parts of this hidden, for example in most card games a player can see their own hand, but not that of the other players. In a wargame the most common form of Imperfect Information is `Fog of War' (FoW), in which a player can only see their own units and those of the enemy within line of sight and range. 
Other forms of imperfect information are possible, for example a player may not have full knowledge of the capabilities of enemy units at the start of a scenario, even when the units are visible.
For the games under consideration the primary aspect of Observability that is relevant is FoW, as summarised in Table~\ref{table:Observability}. 
In this regard they split straightforwardly into Perfect Information games, and Imperfect Information ones with Fog of War.

Hidden information through FoW fundamentally changes some of the challenges faced compared to perfect information games such as Go. We can now only observe part of the full State Space, and formally therefore have to track all possible states that the unobservable parts of the State Space \emph{could} be in. This expands the complexity of the problem, and the most theoretically rigorous analysis of this expansion (in Partially Observable Markov Decision Processes, or POMDPs) is only tractable in small, toy scenarios~\cite{Gmytrasiewicz_Doshi_2005, Ross_Pineau_Chaib-draa_Kreitmann_2011}.
More realistic algorithms that work in large-scale POMDPs usually sample from some estimate of the unknown State Space and are reviewed in more detail in Section~\ref{sect:POMDP}.

\subsection{Observation Space}\label{sect:ObservationSpace}
In a Perfect Information game the Observation Space is by definition the same as the State Space, as this is fully observable.
In an Imperfect Information game the Observation Space is a strict subset of the State Space. For example, where we have Fog of War we can only see some hexes and the full state of our own units.

This should not be interpreted to mean that a smaller Observation Space makes a game easier, and the smaller the Observation Space compared to the State Space, the harder the game in general. Refusing to look at a Chess board when playing would simplify the Observation Space to a single state, and be unlikely to improve one's play. 

The essential categorization of games is covered in Table~\ref{table:Observability}. The third line in this table emphasizes that the mode of Observation can vary:
\begin{itemize}
    \item {Unit-based visibility. This is most standard, with a player able to see anything that any of their units can see.}
    \item {Camera-based visibility. In Starcraft II this is an additional constraint important in a real-time game. A player must physically navigate the camera view to the area they wish to look at, and then they still only see what is visible to their units.}
\end{itemize}
The relevance of the additional camera-based limitation largely depends on the pace of the game. It is akin to a wargame that models the reality of a commander's restricted attention bandwidth when rapid decisions are needed. In a more relaxed turn-based game, or where the passing of time is slow, then it is not relevant.
In wargames the camera-based limitation is not relevant as this approach to modelling attention bandwidth is primarily an artifact of the commercial RTS game genre.

In DOTA 2 each player controls a single unit (hero), and can sees the world from this unit's perspective. However the player also has access to a mini-map that shows the world with all units visible to any other player on the same team. This subtlety does not affect the data in Table~\ref{table:Observability}.

\subsection{Information flow}\label{sect:InfoFlow}
A feature of wargames that is missing from all of the example games here, including Flashpoint and CMANO, is the role of a restricted information flow. In the games of Table~\ref{SummaryTable}, information is presented to the player by visibility of units on a map. In the case of FoW the player (commander) can see anything that their units can see. In physical wargames, as in real combat situations, this is frequently not the case.
Instead the commander is reliant on reports back from units on the ground, which have a restricted informational bandwidth in what they can transmit. A unit that comes under unexpected enemy fire would report this as a priority with perhaps an estimate of the opposing force composition, and not the complete specification of all visible units.

This feature is important in physically-moderated wargames, but is not currently common in 
computer-moderated ones. 
Computer-moderated games are required to incorporate restrictions in Information Flow from commander to units, modelling command and control constraints. 
Information Flow restrictions from units back to commander is anticipated to be a requirement for future wargames.

\subsection{Stochasticity}\label{sect:Stochasticity}
\begin{table}[]
\begin{tabular}[t]{|l|l|l|l|l|l|l|}
\hline 
\textbf{CMANO} & \textbf{Flashpoint} & \textbf{Chess/Go}  & \textbf{Atari/} & \textbf{Starcraft II}    & \textbf{MicroRTS} & \textbf{MuJoCo}\\ 
 &  &   &  \textbf{GVGAI*} & \textbf{+ Dota2}  &  & \\ 
\hline 
Stochastic & Stochastic & Determin. & Determin. & Stochastic & Determin. & Stochastic \\
\hline
\end{tabular}
\caption{Stochasticity categorization. Environments are either Stochastic (have random event outcomes), or Deterministic.  GVGAI has a mixture of deterministic games and stochastic games.}
\label{table:Stochasticity}
\end{table}
This refers to whether game is deterministic or has random (stochastic) elements.  In a deterministic game the same action 
in a given state always leads to the same next state.  In a stochastic game this will lead to a distribution
of possible next states. 
This is closely linked to Branching Factor (Section ~\ref{sect:Branching}), which dramatically increases for stochastic games due to the increase in possible outcomes from a single action, and is a key reason that the minimax search techniques that achieved world championship level play in Chess are less suitable for wargames. These can be converted to an `Expectiminimax' algorithm, but at the cost of more computation due to the branching factor, and a reduced ability to prune the search tree~\cite{russell2016artificial}.
A more common approach is to use Monte Carlo techniques to randomise the outcomes in `rollout' simulations used in both Monte Carlo Tree Search and in the generation of data for Deep RL techniques.

Deterministic single player games are susceptible to over-fitting: agents may learn
a successful sequence of actions, rather than learning how to play the game in a general way. An example of this is learning the precise route for a Pac-Man level given deterministic search patterns by the ghosts; this does not help the agent in the next level.  Environments such as Atari/GVGAI\footnote{The deterministic single-player subset of GVGAI games} and MuJoCo
are susceptible to this kind of over-fitting in RL techniques, though that has been mitigated by adding randomness in the interface between the game engine and the agent~\cite{Haarnoja_Zhou_Abbeel_Levine_2018, Fujimoto_vanHoof_Meger_2018}.

Classic forms of stochasticity in wargames are the uncertainty of combat outcome, or spotting probability of enemy units, minefields etc, which in both physical and computerised versions are decided by dice-rolling~\cite{peterson2012playing}.
This stochasticity introduces the possibility of low-probability but high-impact events in any single run of a game. For example, spotting a well-camouflaged tank column at distance may enable them to be neutralised at low cost, giving a very different outcome in, say, 5\% of games. 
This is a problem for physical wargames, or ones with humans-in-the-loop, as this constrains the number of games that can be run to (usually) low single-figures. For this reason identification of these events is key to determine whether to branch the game (see Section~\ref{sect:wargameTypes}) so that it is representative of the median expected outcome. This is less of an issue in fully computerised wargames with an AI on both sides, as if thousands of games are run, then these low-probability events are representatively sampled and averaged out.

For wargames, there are other sources of uncertainty.  Since they are inherently multiplayer,
the actions of the opponent (whether human or AI) are normally unpredictable, and partial
observability also has similar effects. 
This uncertainty of opponent action (especially with an adversarial opponent) is covered in Section ~\ref{sect:Opponent}, and is not formally `random', but simply not known. If the policy of the opponent were known perfectly, then this might still be random if playing a mixed strategy deliberately, for example in Rock-Paper-Scissors in which a known opponent might play each option with equal random probability.
In practise many game engines may model an AI opponent using an element of randomness, representing the real-world unknown actions they would take. The explicit use of randomness in simulations can provide an efficient way to model a number of types of uncertainty.

For the benchmark game environments, MicroRTS uses deterministic combat, with each unit doing a fixed amount of damage, while CMANO, Flashpoint, DOTA2 and Starcraft II have random damage. CMANO and Flashpoint additionally have random chances for spotting enemy units, affecting the Observation Space.

An interesting way to model stochastic games is taken in the OpenSpiel framework (see Section~\ref{sect:Platforms}).  For most OpenSpiel games any randomness is modelled as the actions of a special random player.  The actions of the random player are recorded, just as they are for any other player.  From the root state of a game, the sequence of actions taken then uniquely defines the resulting game state.  

\subsection{Win Conditions}\label{sect:WinConditions}
\begin{table}[]
\begin{tabular}[t]{|l|l|l|l|l|l|l|}
\hline 
\textbf{CMANO} & \textbf{Flashpoint} & \textbf{Chess/Go}  & \textbf{Atari/} & \textbf{Starcraft II}    & \textbf{MicroRTS} & \textbf{MuJoCo}\\ 
 &  &   &  \textbf{GVGAI} & \textbf{+ DOTA2}  &  & \\ 
\hline 
VP Scale & VP Scale & Win-Lose & VP Scale & Win-Lose & Win-Lose & VP Scale \\
\hline
\end{tabular}
\caption{Win Condition categorization. Environments are either Win-Lose, or have a scale of victory points (VP).}
\label{table:WinConditions}
\end{table}
Games range from simple, clearly defined and unchanging win conditions (e.g. Chess, Go) to 
more complex conditions, which in wargames may be asymmetric, and could potentially change
during the course of a campaign.  In a wargame the same scenario could be played with
different win conditions for each player in order to provide different challenges.

In most games to date, the win conditions are single-objective with binary (or ternary if a game can end in a draw / stalemate) outcomes (e.g. in chess, the aim
is to get the opponent in Checkmate), even though they can be achieved in a multitude of ways.
In this case a player either wins or loses (or draws).

This can be supplemented by a numeric score (or Victory Points) of some sort. Many Atari/GVGAI games have both a win-lose condition combined with a score, for example a treasure hunt game may be `won' when the player reaches the exit before being eaten by a spider, but their score is determined by the number of gold coins they pick up en route.

In a similar way wargames can be multi-objective, where the aim is to achieve a particular outcome
while minimising casualties and economic cost. These are often reformulated, as in Flashpoint,
to a numeric scale by assigning victory points to every feature of the outcome, although
this mapping may be unsatisfactory and can lead to unexpected (and implausible) AI behaviour if the balance of the reward is mis-specified. Biasing the AI in this way can also discourage exploration away from pre-conceived strategies embedded implicitly in the victory points scale.

An alternate, multi-objective, approach is not to weight the different parts of the `score', but seek a range of different policies that are `optimal' in different ways~\cite{Deb_Agrawal_Pratap_Meyarivan_2000, Perez-Liebana_Mostaghim_Lucas_2016}. For example one policy might minimise casualties and fuel usage, while another optimises the number of enemy forces degraded with less concern for the associated cost. This can be a useful way of exploring the space of viable strategies. The multi-objective approach is covered in more detail in Section~\ref{sect:Exploration}.

Of the benchmark games the RTS and Dota2 games are Win-Lose, and differ in this respect to wargames, while some Atari games are closer given the presence of a running score as well as a win-lose condition. MuJoCo is rated as a victory point scale as success is measured by how far and fast an agent learns to move. 

\subsection{Reward Sparsity}\label{sect:RewardSparsity}

This refers to the inherent rewards given \emph{during} a game.  For classic games such as chess, the only inherent reward is at the end of a game when a player wins, loses or draws.  This is closely linked to the Win Conditions of a game in Table~\ref{table:WinConditions}. A game with a binary Win-Lose condition almost by definition also has a sparse reward, while one with a VP Scale of some type provides inherent rewards (VPs) as the game progresses.

Most RTS and wargames have natural features to provide anytime score updates, for example by penalising friendly casualties and loss of assets and incentivising territory gain.
However, these natural heuristics can be misleading in the short-term with immediate losses required (such as sending units into combat to take casualties) as an investment for a longer-term payoff (taking a strategic location). In this sense they have elements of `deceptive' games, in which following the immediate reward signal is counter-productive to final victory~\cite{Anderson_Stephenson_Togelius_Salge_Levine_Renz_2018}.
Hence, methods to promote exploration such as intrinsic motivation and curriculum learning can be very relevant when a key desired outcome is exploration of possible tactics and doctrines, as in Concept Development wargames.

Hence, while wargames, like RTS, do have a short-term point score many of the AI techniques developed to cope with sparse rewards are important. These are looked at in more detail in Section~\ref{sect:RelativeRS}.


\subsection{Active/Adversarial Opponents}\label{sect:Opponent}
\begin{table}[]
\begin{tabular}[t]{|l|l|l|l|l|l|l|}
\hline 
\textbf{CMANO} & \textbf{Flashpoint} & \textbf{Chess/Go}  & \textbf{Atari/} & \textbf{Starcraft II}    & \textbf{MicroRTS} & \textbf{MuJoCo}\\ 
 &  &   &  \textbf{GVGAI} & \textbf{+ Dota2}  &  & \\ 
\hline 
Adaptive & Adaptive & Adaptive & Non-Adapt. & Adaptive & Adaptive & Non-Adapt. \\
Adversarial & Adversarial & Adversarial &  & Adversarial & Adversarial &  \\
\hline
\end{tabular}
\caption{Opponent categorization. Opponents may be Adaptive to a players actions and may or may not also be Adversarial to them.}
\label{table:Opponent}
\end{table}
An essential component of a wargame is one or more opponents. In this they contrast with many Atari games in which a single-player fights against the environment, as in Space Invaders, Pac-Man or Asteroids. In these, the enemies are scripted units that do not change behaviour or adapt to player actions. Other Atari games have enemies that do adapt, for example the ghosts in Ms Pac-Man will change direction to move towards a visible player, however these remain simple reactive scripts and are not `Adversarial' in terms of planning a strategy to win the game.
This contrasts with Chess, Go, RTS games or wargames, in which there is a clearly defined opposing force that makes proactive decisions to achieve objectives and adapt to the player's actions. 

All the benchmark games considered are one or two player games. Some wargames can have more than two sides, for example a neutral Brown or White supranational force, or a Pink side that may, depending on what happens, become actively allied to Blue or Red. A large Planned Force Test will have multiple autonomous players on each side, with restricted communication between them even if they are collaborating on a joint goal.
However, the computerised wargames directly relevant to AI control are Red vs Blue 2-player situations, and here the essential requirement to deal with large numbers of units controlled by a single player is covered in detail in Section~\ref{sect:StateSpace}.

There is also at this stage relatively little academic work that addresses the unique aspects of wargames with more than two interacting players, especially where communication can occur. This can introduce `king-making' in which a third player can decide which of the others achieves a victory objective, and negotiation between players can become vital to win\cite{elias2012characteristics}.

\subsection{Scenario Variability}\label{sect:ScenarioVar}
\begin{table}[]
\begin{tabular}[t]{|l|l|l|l|l|l|l|l|}
\hline 
\textbf{CMANO} & \textbf{Flashpoint} & \textbf{Chess}  & \textbf{Atari/} & \textbf{Starcraft II}  & \textbf{Dota2}  & \textbf{$\mu$RTS} & \textbf{MuJoCo}\\ 
 &  & \textbf{/Go}  &  \textbf{GVGAI} & &  &  & \\ 
\hline 
Scenario & Scenario & Fixed & Scenario & Scenario & Scenario & Scenario & Fixed \\
\hline
\end{tabular}
\caption{Scenario categorization.}
\label{table:Scenario}
\end{table}
In most classic board games the start state is always the same, whereas in wargames
the same underlying rules may be played out over a wide range of initial scenarios. 
In Table~\ref{table:Scenario} games can relatively cleanly be split into those that have some concept of `scenario' (for Atari games these are different `levels', which use the same underlying rules but with different maps), and those which have a single fixed set up.
In the case of DOTA2 there is little difference between maps compared to RTS and wargames, however a similar level of variability comes in due to the combinations of different heroes that form both teams; and with a slight stretch we can consider these different opposing teams as different `scenarios'.

The biggest risk of using techniques with a single fixed set up is of over-fitting to the map. This is not an issue in Chess or Go, in which we want to overfit; learning a policy able to generalise well to a 7x7 board, or with half the number of starting pawns would be wasteful. 
In wargames, as in RTS games generally, we are only interested in AI techniques that can generalise to scenario variations. 
As a note of caution, some papers in the literature use games that support scenarios/levels, but only report results that are trained and tested on one specific set up.
Planning techniques in general are more robust in the case of scenario variability than methods without a planning component, such as \emph{vanilla} RL.

In particular, both MicroRTS and GVGAI competitions put the trained agent through multiple maps, at least some of which are unseen prior to the competition. It is noticeable that the winning entries in both competitions (see~\cite{Perez-Liebana_Samothrakis_Togelius_Schaul_Lucas_Couetoux_Lee_Lim_Thompson_2016, Ontanon_Barriga_Silva_Moraes_Lelis_2018} for a summary of these) use planning and search-based methods with little direct RL, even though the decision time for each move is only in the range 40-100 milliseconds on a single processor. The MicroRTS winners tend to use a two-level hierarchical approach to select amongst scripts to use for different units (these are discussed in more detail in Section~\ref{sect:unitTechniques}). 

The headline work from DeepMind and OpenAI on Starcraft II and Dota2 using Deep RL does support multiple `scenarios' in the sense of having different heroes, maps and opposing races. However these are all available at the start and included in the training process. There is little work showing good performance with Deep RL on unseen environments, although this is an area of active research interest in terms of transfer-learning. 
This does not mean Deep RL is not useful in the context of wargames, but is not suitable for completely unseen scenarios. Any substantive change to a scenario will require some training of a Deep RL model for good performance.
The issues around this overfitting in Deep RL are discussed in more detail, along with mitigation approaches, in Sections~\ref{sec:continual_learning} and~\ref{sect:overfit}.

\subsection{Objective Variability}\label{sect:ObjectiveVar}
In a wargame the objective that a player is striving for may change as the game progresses. For example due to changes or setbacks on other fronts, a player attempting to take a strategic location may be ordered to fall back and reinforce elsewhere. This can cause the `reward' function or win condition to change drastically.
Wargames such as CMANO come with a large number of scenarios,
and these can potentially be played out with a number of different
objectives for each player.  This presents a significant generalisation
problem for Deep RL algorithms, which are normally trained with
a specific objective in mind, and transferring what they learn
during training on one objective to then act in the pursuit of a different objective is likely to remain a research challenge for some time.

If this variability is incorporated during training, then this is less of an problem. This could be via a multi-objective approach, or incorporating the possible objectives in a VP-scale (see section~\ref{sect:WinConditions}). 
Planning/search methods with a forward model will be better at adjusting to changing objectives if this is needed.

None of the considered benchmarks really incorporate any variation of game objectives in this way. 

\subsection{Player Asymmetry}\label{sect:PlayerAsymmetry}
\begin{table}[]
\begin{tabular}[t]{|l|l|l|l|l|l|l|}
\hline 
\textbf{CMANO} & \textbf{Flashpoint} & \textbf{Chess/Go}  & \textbf{Atari/} & \textbf{Starcraft II}    & \textbf{MicroRTS} & \textbf{MuJoCo}\\ 
 &  &   &  \textbf{GVGAI} & \textbf{+ DOTA 2}  &  & \\ 
\hline 
Asymmetric & Asymmetric & Symmetric & - & Asymmetric & Symmetric & - \\
\hline
\end{tabular}
\caption{Player asymmetry categorisation. Atari and MuJoCo are 1-player environments.}
\label{table:Asymmetry}
\end{table}
This refers to the asymmetry in player capabilities and roles.  For example, although chess may have a slight first-player advantage, the action space is the same for each player, and the game is well balanced and largely symmetric.  On the other hand, in the Ms Pac-Man versus ghosts competition, AI agents could play as Pac-Man, or play as the entire team of ghosts, hence the objectives and the actions spaces are both asymmetric.

Wargames are in general highly asymmetric, with differences in capabilities, unit compositions, starting positions and ultimate objectives between the two sides. Even the RTS games in Table~\ref{table:Asymmetry} are relatively symmetric compared to wargames; in Starcraft II there are four possible factions with specific units, but each game starts with a similar set up and both sides have the same objective to destroy the other. In MicroRTS the same units and technology are shared by both sides. DOTA 2 might have very different teams of 5 heroes on each side, but the map is essentially symmetric, as is the goal of each side to destroy the other's base.

High levels of asymmetry prevent easy use of some AI techniques, especially those that rely on self-play in RL. In this approach the AI can bootstrap its playing ability by using itself, or a slightly older version of itself, to play the opponent. However, the policy that the Blue side is learning is not \textit{a priori} going to work very well if used to model the actions of an asymmetric Red side.

\subsection{Features specific to wargames}\label{sect:wargameSpecific}
This section discusses some requirements of practical wargames not in directly linked to the game-play itself, but to the underlying purpose of running the wargame, and the efficient generation of desired outcomes.
\subsubsection{Instrumentation}
A frequent desired outcome of wargames is to understand key decision points in a game rather than just the win-lose result. The end-user is the wargame analyst.
For example, a rare event such as the early sinking of Blue's only aircraft carrier due to a lucky missile hit may mean that playing out the rest of the scenario is of limited value. What is important to know is that there is a risk of this happening, perhaps even quantified that this occurs in 2-5\% of games. 
Finding this important insight can require excessive data mining and be easy to miss. Wargames hence require rich instrumentation of the trajectory that an individual game takes to facilitate anomaly detection of this sort of situation and reduce the work required to understand the location of common branch points.
\subsubsection{Moderation}
Computer-moderated wargames always have options for human intervention to adjust the state of play at any point and/or back-track to a previous point. This is required when one side is being played by humans to make efficient use of their time. If a low-probability high-impact event occurs, such as the aircraft carrier loss of the previous example, then the decision may be made to branch at that point, log the result, and then continue play in a counterfactual universe.
This need for human moderation is less of an issue if all sides are AI-controlled.
\subsubsection{Player Constraints}
One issue with human players in wargames is a tendency to `play the game' rather then `play the problem'. Examples are the `last turn effect' in which players may gamble everything on a chance of victory even if in the real world this would leave them heavily exposed to future enemy action~\cite{rubel2006epistemology}.
This is an issue to which AI agents are particularly prone without careful specification of the reward signal they use as outlined in Section~\ref{sect:WinConditions}.

A related requirement is often for players to be `credible' in terms of following operational constraints, such as orders from further up the command hierarchy or standard operating procedures for the side in question. This only applies in some types of wargames; in Concept Development the objective is to deliberately explore diverse and unexpected strategies with new equipment or capabilities.
We discuss useful techniques for this later in Section~\ref{sect:Exploration}.

One issue with practical wargames is that valuable time of participating human experts can be taken up with micro-management of units. This detracts from the central aim of the wargame to analyse/train human decision-making. A helpful tool in some wargames would be to have an AI control units once given a high-level order. It is essential that this is realistic as naive use of A* pathfinding and simple scripted AI has shown problems with units moving directly from A to B without paying attention to available cover or the unexpected appearance of enemy forces.

\vfill
\pagebreak

\section{Recent Progress in Deep Reinforcement Learning}\label{sect:DeepRL}

An agent in reinforcement learning tries to solve a sequential decision making problem under uncertainty through a trial-and-error learning approach with respect to a reward signal \cite{sutton1998introduction}. The approach can often be described as follows: the agent receives a state $s_t$ at timestep $t$ from the environment and based on this state, samples an action $a_t$ from its policy $\pi(a_t|s_t)$. After performing the chosen action, the environment transitions to the next state $s_{t+1}$ and the agent receives a scalar reward $r_{t+1}$. The main objective in an RL setting is to optimize the agent's policy in order to maximise the sum of rewards.

A particularly popular early RL method was Q-learning, which tries to learn an optimal state-action value function $Q(s, a)$ describing the best action $a$ to take in a particular state $s$. However, a challenge with this original tabular version of Q-learning is that the size of the table quickly becomes infeasible when the number of different states in the environment grows. Additionally, learning in such a sparse table would not be very efficient.

A breakthrough introduced in 2015 by DeepMind was a method called Deep Q-Network (DQN), in which $Q(s, a)$ is represented by a  deep neural network that learns low-level representations of high-level states \cite{atari}. The DQN approach was the first approach to being able to master many Atari games at a human level from pixels alone. 

Since then, many different deep reinforcement learning approaches combining the representational power of neural networks with reinforcement learning have shown impressive results in a variety of different domains.
%
%
Two of the most impressive applications of Deep RL in games are  DeepMind's work on StarCraft II \cite{vinyals2019grandmaster} and OpenAI's bot for Dota2 \cite{berner2019dota}. In the following we will review these approaches in terms of their computational and also engineering efforts.  Both StarCraft II and Dota2 not only have a very high action space but many aspects are continuous (see Sections~\ref{sect:StateSpace} and~\ref{sect:ActionSpace}). Additionally, we will review other recent trends in RL that are particular important for wargaming, such as being able to continually learn, to produce explainable actions, or hierarchical deep RL methods.  


\subsection{Dota2}
One of the potentially most relevant applications of deep RL to games, in regards to the challenges faced in typical wargames, is OpenAI's OpenFive bot \cite{berner2019dota}. This was the first AI system to defeat world champions at an e-sport game. In contrast to DeepMind's work on AlphaStar that learns from pixels, OpenFive learns from pre-processed information on enemy and unit locations. This setup is likely more relevant for an AI that is tasked with playing wargames such as CMANO or Flashpoint. 

Dota2 is a multi-player real-time strategy game with complex rules that have evolved over several years. The game is played on a square map with two teams of five players each competing against each other. Each player controls one of the five team members, which is a hero unit with special abilities. Players can gather resources through NPC units called ``creeps" and use those to increase the power of their hero by purchasing new items and improving their abilities. 

Dota2 is challenging for RL systems since it includes imperfect information (not all parts of the map are visible at the same time; see Section~\ref{sect:Observability}), long time horizons, and continuous state and action spaces. Games last for approximately 45 minutes and the state of the world is only partially observable, limited to the areas near units and buildings. 

\subsubsection{Neural Network architecture}

\begin{figure}[ht]
\centering
\includegraphics[width=0.9\textwidth]{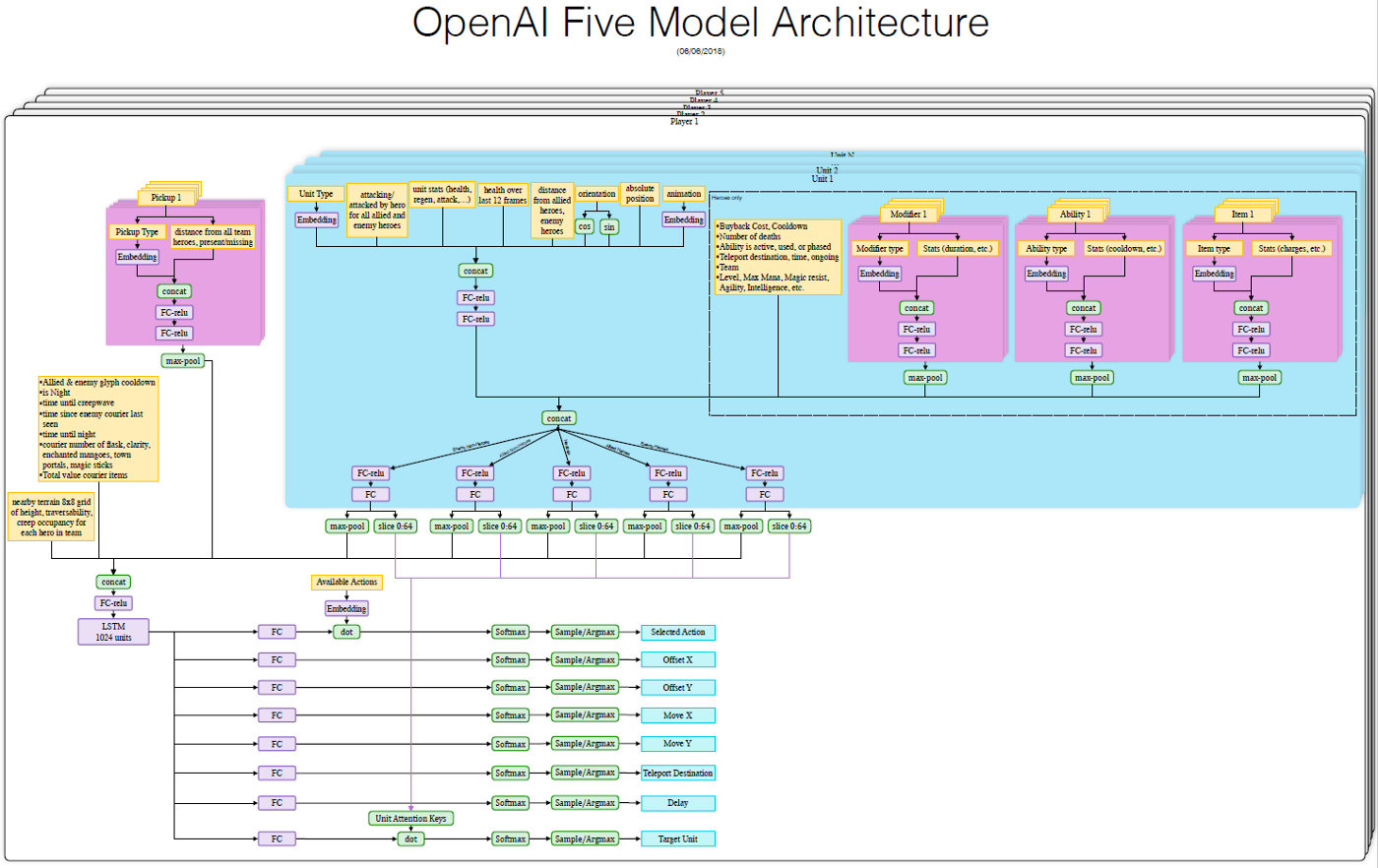}
\caption{Neural architecture for OpenAI Five \cite{berner2019dota}.}
\label{fig:open_five}
\end{figure}

The OpenFive neural network architecture that controls each agent in the game  observes around 16,000 inputs each time step (Figure~\ref{fig:open_five}). The majority of these input observations are per-unit observations for each of the 189 units in the game (e.g.\ 5 heroes, 21 buildings, etc). The action space is divided into primary actions that the agent can take (e.g.\ move, attack) and actions are then  parameterised with additional neural network outputs. Availability of actions was checked through hand-designed scripts.  In total, the action space had 1,837,080 dimensions. 

Each of the five agents is controlled by replicas of the same network. Observations are shared across all members of the team (the network gets as input an indicator which unit it is controlling). The employed network is a recurrent neural network with approximately 159 million parameters, mainly consisting of a single-layer 4096-unit LSTM.

As mentioned above, instead of working from the screen pixels directly, information about opponent positions etc, which would also be available to a human player, is pre-processed and given to the neural network directly. 
This way the system is focused on high-level decision making and strategic planning instead of visual information processing. Such as system thus aligns well with the setup likely encountered in wargames. Some game mechanics were controlled by hand-designed scripts, such as the order in which heroes purchase items. Additionally, while humans need to click on certain parts of the map to view it and obtain information, this is instantaneously available to the network at each time-step.
This means the AI arguably has an advantage over human-players in reducing the number of clicks needed to get information, however this `click-to-view' is an artefact of the real-time game genre, and the key point for wargames is that the AI does not have access to information unavailable to human players.


\subsubsection{Training Regimen}
The network was trained with a hand-designed reward function that not only rewarded the agent for winning the game (which would be a very sparse reward) but includes additional signals such as if another character died, or which resources were collected. This reward function was constructed by people familiar with Dota 2, and according to the paper, not altered much during the project. Instead of manually designing reward functions, these can also be directly learned from human preferences, which we explain in more detail in Section~\ref{sect:imitation}. 

The particular RL training algorithm was Proximal Policy Optimization (PPO) \cite{schulman2017proximal}, which is a state-of-the-art policy optimization method. The system is trained through \emph{self-play}, in which the agent plays 80\% of games against itself and 20\% against an older version of its policy. 

The system implements a complicated training architecture with both GPU and CPU pipelines that handle the game rollouts, to reach the large number of samples needed for training: ``After ten months of training using 770$\pm$50 PetaFlops/s days of compute, it defeated the Dota 2 world champions
in a best-of-three match and 99.4\% of human players during a multi-day online showcase." \cite{berner2019dota}.

In conclusion, the OpenFive agent uses a very computationally expensive approach that required extensive fine-tuning of its reward function. Thus, in its current form, it is likely not the best approach for wargaming. 


\subsection{AlphaStar}


Even more so than Dota2, StarCraft II is incredibly complex. Instead of five units, players in StarCraft control hundreds of units. Additionally players need to balance short and long-term goals. Therefore, AlphaStar \cite{vinyals2019grandmaster} uses a more complicated neural architecture than the OpenFive Dota agent, combining multiple recent deep learning advancements, such as a neural network transformer \cite{vaswani2017attention}, LSTM core \cite{hochreiter1997long}, pointer network \cite{vinyals2015pointer}, and centralised value baseline \cite{foerster2018counterfactual}. 

While an RL approach by itself was able to learn to play the game at a high-level, it is important to note that Grandmaster level in this game was only reached when starting training with human demonstrations. Training in a purely supervised way, the system was already able to reach gold level play. Training the agent further through RL, allowed  AlphaStar to be above 99.8\% of officially ranked human players, reaching Grandmaster status. In Section~\ref{sect:imitation} we will go into more detail on how human demonstration and related methods can bias RL systems towards human-like play.

The previously mentioned AlphaStar tournament-like training setup (also called the AlphaStar league) points to the fact that it is incredibly difficult to make sure agents do not overfit to a particular playstyle or opponent, which would allow them to be easily exploited. 

Interestingly, AlphaStar did not employ any search-based planning method. Instead, the LSTM-based network learned to do some form of planning by itself. However, training this very complex system comes at a cost: ``The AlphaStar league was run for 14 days, using 16 TPUs for each agent. During training, each agent experienced up to 200 years of real-time StarCraft play."


\subsection{Frameworks for scaling up training of deep RL agents}
\label{sec:deep_rl_frameworks}

While AlphaStar and Dota 2 relied on their own engineered  training system, a few frameworks have since been released that have the potential to reduce the engineering efforts required for large-scale distributed AI computing. Therefore, these systems are likely of relevance for improving capabilities in the area of AI and wargaming. We identified one particular framework, that could significantly reduce such engineering efforts.

Fiber\footnote{https://uber.github.io/fiber/introduction/} is a framework recently released by Uber to more easily scale machine learning approaches such as reinforcement learning and evolutionary optimization to many different machines in a compute cluster~\cite{wang2019poet}. Fiber tries to address common issues with scaling RL up to a large number of computers, such as (1) time it takes to adapt code that runs locally to work on a cluster of computers, (2) dynamic scaling to the amount of available resources, (3) error handling when running jobs on a cluster, and (3) high learning costs when learning a new API for distributed computation. 

Fiber seems easy to learn, since it uses the same API as Python multiprocessing (which many RL approaches are already based on) and applications can be run the same way they are normally run without extra deployment steps. The framework is also supposed to work well with existing machine learning frameworks and algorithms. 

Other existing frameworks include PyTorch-based Horizon \cite{gauci2018horizon}, which allow end-to-end RL training but seems less flexible than Fiber to quickly try out new ideas. Dopamine is another tensorflow-based system \cite{castro2018dopamine} that is flexible to use but does not directly support distributed training.

\subsection{Automatically Learning Forward Models for Planning}
\label{sec:learning_fm}

Wargames would benefit from agents that can do long-term strategic planning. While these abilities can in principle emerge in an LSTM-based neural network that learns to play (such as AlphaStar), this approach is in generally computationally expensive and requires very long training times.

An alternative approach that has gained significant attention recently is to learn a model of the world through machine learning \cite{ha2018world} and then use that model for a planning algorithm \cite{schrittwieser2019mastering,hafner2019dream,lucas2019local,ha2018world}. This approach tries to combine the benefits of high-performing planning algorithms \cite{Browne_Powley_Whitehouse_Lucas_Cowling_Rohlfshagen_Tavener_Perez_Samothrakis_Colton_2012} with breakthroughs in model-free RL. 
For example, the muZero approach developed by DeepMind~\cite{schrittwieser2019mastering}, first learns a model of the environment's dynamics and then uses Monte-Carlo Tree Search at each timestep in the learned forward model to find the next action to execute in the actual environment. The benefit of this approach is that planning can happen in a smaller latent space, in which it is not trained to predict the next pixels but instead the reward, the action-selection policy, and
the value function. These functions are trained through backpropagation based on trajectories sampled from a replay buffer.  Using a small latent space for planning is computationally faster and only incorporates the information the system learns is needed to make effective decisions; for example by ignoring wall textures or pixel-animations of sprites, both of which take up a lot of 'information' in the raw pixel input and do not need to be included in a useful forward model.
%
Their approach showed new state-of-the-art performance in both the visually rich domain of Atari games, and the board games Go, Chess, and Shogi.

However, computational requirements are rather heavy for this approach. For the three board games, they used 16 TPUs for training and 1,000 TPUs for selfplay. Agents were trained for 1 million training steps using 800 simulation steps per move. For the Atari games, 8 TPUs were used for training and 32 TPUs for selfplay. In 2019, cost of a a cloud TPU v3 device was at \$8 US per hour. The time for training the Atari agents was 12 hours (for the final model, not taking into account models trained during experimentation) and is likely much higher for Go. We estimate one training run on Atari  costs around 40 TPUs * 12 * 8 = \$3,840 US and running 36 hours training for the board games would be 1,000 TPUs * 36 * 8 = \$288,000 US.

One thing to note is that algorithms are also becoming more efficient to train, and increasing in efficiency more than Moore's law would predict\footnote{https://openai.com/blog/ai-and-efficiency/}. For example, it now takes 44 times less compute to train a high-performing network for state-of-the-art image recognition than it took in 2012 (Moore's law would have only predicated a 11x cost improvement). Similar trends can be observed in other domains. For example, AlphaZero required 8x less compute than AlphaGoZero. 
The increasing commoditisation of GPU/TPU facilities in the cloud have also contributed to decreasing the overall cost of computation over time.

\subsection{Explainable AI}\label{sect:ExplainableAI}
To best inform decisions on future force structures based on the results from the wargaming experiments, it would be ideal if the algorithm could explain its decision. For example, did the AI attack the ship because it believed it to have a particular weapon on board and would its decision have been different, if that wasn't the case.

While methods based on deep neural networks are often considered to be black boxes that are difficult to analyse, recent advances have allowed more insights into the decision making process of neural networks. These techniques can be roughly divided into approaches that
aim to (a) visualize features, and (b) elucidate the relationship between neurons. Visualizing hidden layers’ features can give insights into what network inputs would cause a certain reaction in an internal neuron or in the network output. 

\begin{figure}
\centering
\includegraphics[width=1.0\textwidth]{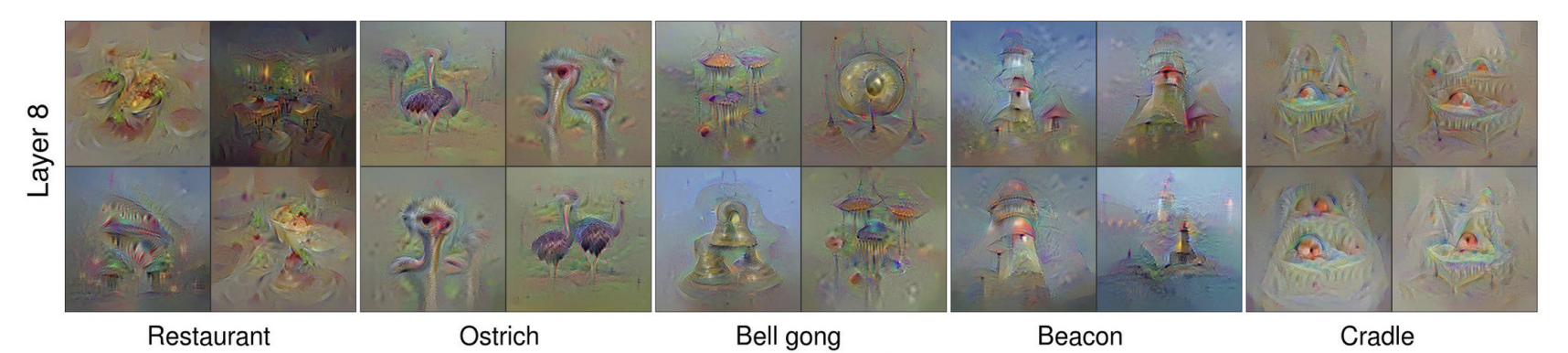}
\caption{Activation maximization \cite{nguyen2016multifaceted}. These show the input (in this case, an image) that would maximally activate a specific neuron, to show what patterns the neural network is looking for.}
\label{fig:activation_max}
\end{figure}

In contrast to optimizing the network’s weights, as normally done through gradient descent to train the network, one can use the same process to optimize the input to the network that would maximally activate a certain neuron \cite{nguyen2016multifaceted}. These techniques can  help to identify what features certain neurons learned to pay attention to (Figure~\ref{fig:activation_max}). Instead of looking at only single features, more advanced methods look at combinations of features to determine which are important for the network to reach its decision \cite{olah2020zoom}. For example, to identify a car, these methods determined that the network learns to identify windows, car bodies, and wheels; only if all three are present does the network predict a car.  

\begin{figure}[h]
\begin{subfigure}{.5\textwidth}
  \centering
\includegraphics[width=0.9\textwidth]{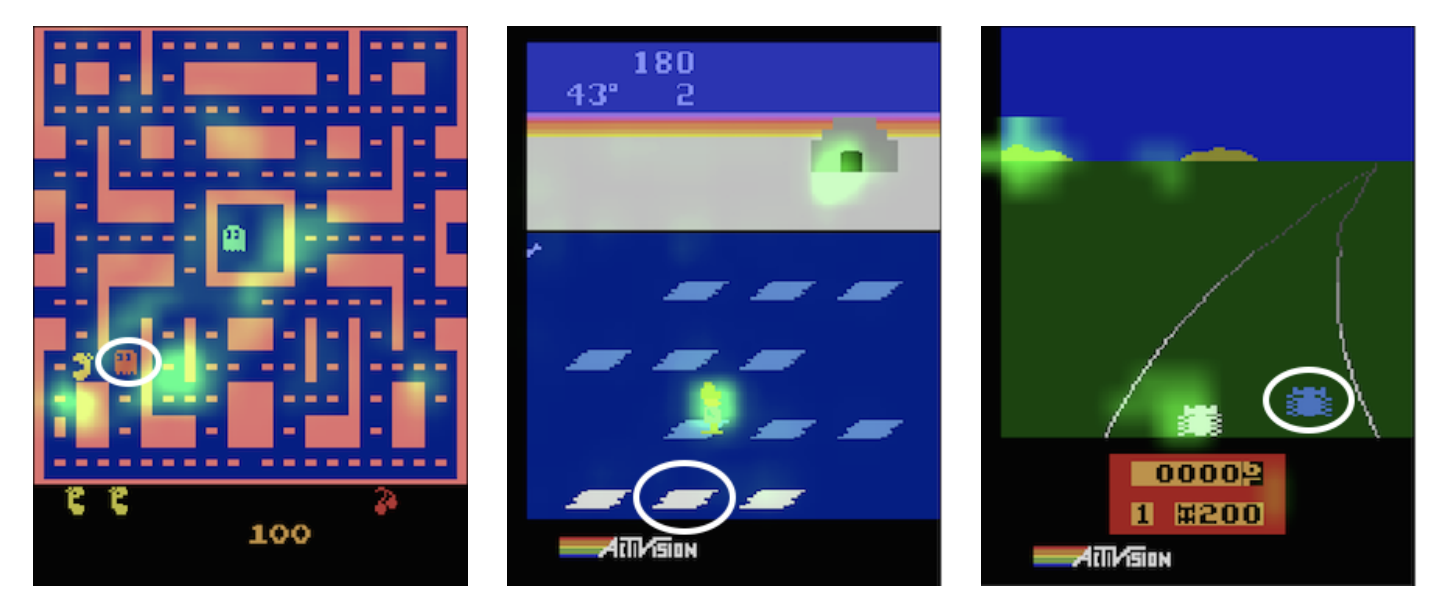}
  \label{fig:sfig1}
\end{subfigure}%
\begin{subfigure}{.5\textwidth}
  \centering
\includegraphics[width=0.9\textwidth]{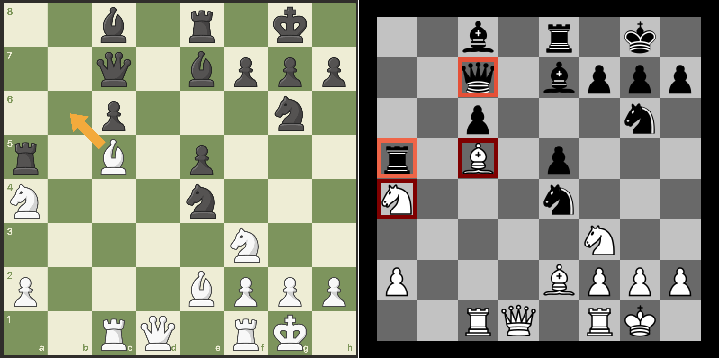}
  \label{fig:sfig2}
\end{subfigure}

\caption{Saliency maps~\cite{greydanus2017visualizing, Gupta_Puri_Verma_Kayastha_Deshmukh_Krishnamurthy_Singh_2020}. The brighter pixels in (a) show the areas of the neural network input that affect the decision (i.e. changing them, changes the decision). In (b) the highlighted pieces are the ones that affect the neural network decision to move the bishop as shown on the right. }
\label{fig:saliency}
\end{figure}

Similarly these approaches can create saliency maps (Figure~\ref{fig:saliency}) that indicate how important each input signal is for the network outputs. Similar approaches cover parts of the image or blur them out and these occupancy-based salience maps can similarly give insights into what types of input are important for the network. While these approaches are typically applied to image-based inputs, they can also work for other input representations~\cite{Gupta_Puri_Verma_Kayastha_Deshmukh_Krishnamurthy_Singh_2020}. 

Additionally, by giving  counterfactual examples to the network (e.g.\ instead of weapon type A, the network now receives weapon type B as input), can give insights into the situations that resulted in the network performing a certain action \cite{myers2020revealing}.

\subsection{Continual adaptation and fast learning}
\label{sec:continual_learning}

While applications of Deep RL have shown impressive results,  they only perform well in situations they have been trained for in advance and are not able to learn new knowledge during execution time. If situations or circumstances change only slightly, current Deep RL-trained AI systems will likely fail to perform well. These issues clearly limit the usage of these machines, which have to be taken offline to be re-trained while relying heavily on human expertise and programming.

In the context of wargaming, a new unit type might be introduced or the strength of existing units could change. One approach would be to just retrain the system from scratch, but depending on the frequency of changes to the game and the computational training demands, this approach could quickly become infeasible. 

Because the code and settings for Dota 2 changed over time, OpenAI developed an approach they called `surgery' that allowed them to resume training. A similar approach could be deployed for wargaming. However, ideally, an approach should be able to continually incorporate new knowledge and adapt to new circumstances during deployment without having to be taken offline for training. 

While current deep learning systems are good at learning a particular task, they still struggle to learn new tasks quickly; meta-learning tries to address this challenge. The idea of meta-learning or \emph{learning to learn} (i.e.\ learning the learning algorithms themselves) has been around since the late 1980s and early 1990s but has recently gathered wider interest.

A family of meta-learning approaches trains a special type of recurrent network (i.e.\ an LSTM) through gradient-based optimization that learns to update another network~\cite{ravi2016optimization, zoph2016neural}. In an approach based on reinforcement learning (RL) an agent learns how to schedule learning rates of another network to accelerate its convergence \cite{fu2016deep}. Typically, such networks are trained on several different tasks and then tested on their ability to learn new tasks. A recent trend in meta-learning is to find good initial weights through gradient-based optimization methods from which adaptation can be performed in a few iterations. This approach was first introduced by \cite{finn2017model} and is called Model-Agnostic Meta-Learning (MAML). 
A similar approach is used in Natural Language Processing (NLP), with pre-trained networks used as a starting point (having been trained on all of wikipedia for example, to gain a solid model of English syntax), and then specialised in some domain, for example by fine-tuning on Hansard to parse political speeches, or medical journals for an application in medicine.



\subsection{Overfitting in RL agents}\label{sect:overfit}
A challenge with employing RL agents for wargames is that pure RL algorithms often overfit to the particular scenario they are trained in, resulting in policies that do not generalize well to related problems or even different instances of the same problem. Even small game modifications can often lead to dramatically reduced performance, demonstrating that these networks learn reactions to particular situations rather than general strategies~\cite{kansky2017schema,zhang2018study}. For example, to train OpenAI Five, several aspects of the game had to be randomized during training (e.g.\ the chosen heroes and the items they purchased),  which was necessary for robust strategies to emerge that can handle different situations arising when playing against human opponents. 

This type of overfitting can also be exploited for adversarial attacks. For example, Gleave et al.~\cite{gleave2019adversarial} showed that an agent acting in an adversarial way (i.e.\ a way unexpected to another agent), can break the policy of humanoid robots trained through RL to perform a variety of different tasks. In one interesting case, a robot goal keeper just falling to the ground, twitching randomly, can totally break the policy of an otherwise well performing goal-shooting robot.

One idea to counter overfitting is to train agents on a a set of progressively harder tasks. For example, it has been shown that while evolving neural networks to drive a simulated car around a particular race track works well, the resulting network has learned only to drive that particular track; but by gradually including more difficult levels in the fitness evaluation, a network can be evolved to drive many tracks well, even hard tracks that could not be learned from scratch~\cite{togelius2006evolving}. 
Essentially the same idea has later been independently invented as curriculum learning~\cite{bengio2009curriculum}. Similar ideas have been formulated within a coevolutionary framework~\cite{brant2017minimal}.

Several machine learning algorithms also gradually scale the difficulty of the problem. Automated curriculum learning includes intelligent sampling of training samples to optimize the learning progress \cite{graves2017automated}. Intelligent task selection through asymmetric self-play with two agents can be used for unsupervised pre-training \cite{sukhbaatar2017intrinsic}. The \textsc{powerplay} algorithm continually searches for new tasks and new problem solvers concurrently \cite{schmidhuber2013powerplay} and in Teacher-Student Curriculum Learning \cite{matiisen2017teacher} the teacher tries to select sub-tasks for which the slope of the learning curve of the student is highest. Reverse curriculum generation automatically generates a curriculum of start states, further and further away from the goal, that adapts to the agent's performance \cite{florensa2017reverse}. 

A promising recent method to combat overfitting in deep RL is to train agents on a large number of different procedurally generated environments \cite{risi2019increasing, justesen2018illuminating}.  In the \emph{progressive procedural content generation} approach \cite{justesen2018illuminating}, an agent is presented with a completely new generated level each episode during reinforcement learning. If the agent is able to solve the level, difficulty of the generated environments is increased. This approach allows agents to solve levels they would normally not be able to solve, and also produces agents that overfit less and generalize better to other levels.

\subsection{Hierarchical RL}\label{sect:HRL}
A particularly relevant form of RL for wargaming is hierarchical RL (HRL). HRL aims to scale up RL by not just training one policy but a hierarchy of policies. In this hierarchy, some actions operate on low level actions while other levels can make use of these simpler actions to perform macro-actions. This type of abstraction greatly reduces the search space of the problem but finding the right level of abstraction remains an important challenge in HRL.

Some of the most popular forms of HRL are `options frameworks' \cite{sutton1999between} and `feudal RL'. In the options framework, actions can take a variable amount of time. More specifically, options consist of an option policy, an initiation set, and a termination set. A particular option is available to take in a certain state $s$, if $s$ is part of that option's initiation set. In that case the policy is executed until it terminates stochastically according to some termination condition. Early work in this area dealt with pre-defined options but more recent work focuses on learning the appropriate options.

\begin{figure}[h]
\centering
\includegraphics[width=0.7\textwidth]{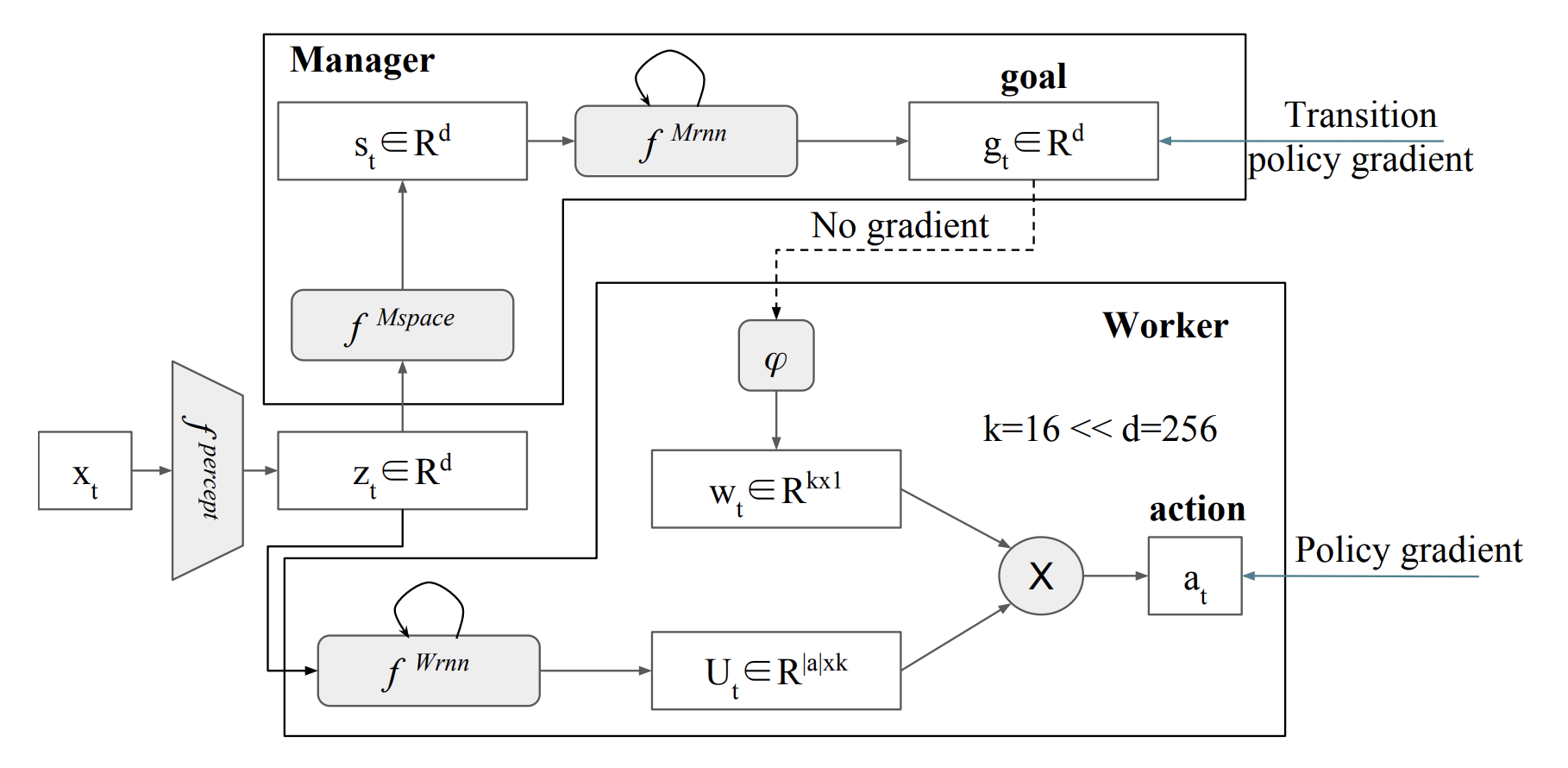}
\caption{FeUdal Network Architecture \cite{ahilan2019feudal}.}
\label{fig:feudal}
\end{figure}

 In feudal RL, levels of hierarchy within an agent communicate via explicit goals but not exactly how these goals should be achieved. More recent work in RL extends the idea of feudal RL to deep neural networks \cite{ahilan2019feudal}. In this work (Figure~\ref{fig:feudal}),  a feudal system is composed of two different neural networks. A manager network sets goals at a lower temporal resolution for a worker operating at a higher temporal resolution. The worker performs primitive actions conditioned on the goal given by the manager network. This approach has shown promise in hard exploration problems such as the Atari game Montezuma’s revenge. While typical RL algorithms only reach very low scores in this game, the FeUdal approach allowed the manager to learn meaningful sub-goals for the worker, such as picking up the key in the environment or reaching the ladder.  
 
 For wargaming, the feudal approach is likely more relevant than the options framework, since it already reflects the typical hierarchical troop structure in combat scenarios.


\subsection{Deep Multi-agent Learning}
OpenFive and AlphaStar present two different forms of deep multi-agent learning. OpenFive can be described as a distributed homogeneous multi-agent system, in which each agent is controlled by clones of the same network but there is no overall manager network. AlphaStar is using a different approach, in which one centralised manager network controls all the available units.  Other approaches such as distributed heterogeneous multi-agent systems train a different neural network for each agent, based on approaches such as 
independent Q-learning (cite). However, it is difficult to scale these  approaches to a large number of agents. Multi-agent learning and HRL methods have also been combined \cite{kumar2017federated}. 

An important factor to consider for wargaming is scalability. Ideally we do not want to retrain the whole system from scratch when a new unit type if introduced, which limits the use of centralised controllers. Heterogeneous multi-agent RL approaches are also divided into approaches in which the environment becomes partially observable from the point of each agent that might not have access to the whole state space. In the context of wargaming agents, we can assume that all agents have access to all information all the time.




\vfill
\pagebreak

\section{Other Game AI techniques}\label{sect:OtherAlgos}
Deep Reinforcement Learning has been behind recent headlines in achieving human or world championship level play in game domains from Atari, Go, Starcraft II and DOTA 2 amongst others. Other AI techniques do exist, and this section briefly summarises the main ones.
There are good reasons for considering techniques other than Deep RL:
\begin{itemize}
    \item \textbf{Complementarity}. Deep RL is often used in conjunction with other techniques. AlphaGo for example uses Deep RL to learn a position evaluation function, which estimates the chance of victory from a given state of the game board. This is then used within Monte Carlo Tree Search (MCTS), a Statistical Forward Planning algorithm, to decide on which move to make.
    \item \textbf{Raw Sensory input}. The Atari, Go and Starcraft II benchmarks all use raw pixels as their input. This stems in part from the underlying research agenda to learn to play without any \textit{a priori} domain knowledge. With this agenda, games are a stepping stone towards Artificial General Intelligence (AGI) that learns from raw sensory input. In a practical application this is not a requirement and we can and should use domain knowledge, existing forward models and available APIs. 
    Deep RL is especially useful in raw sensory input domains, with the `Deep' layers designed to extract the useful features. Convolutional Neural Networks (CNNs) in particular are constructed to build up representations at decreasing levels of granularity, from small pixel groups at the first layer to cat faces at later ones.
    Note though that the Dota2 work used Deep RL without pixel-based inputs; Deep RL is still useful without such constraints but the structure of the networks used will differ somewhat, and may be shallower for optimal performance.
    \item \textbf{Computational efficiency}. The Dota2 final model required 10 months of continuous training across 51,000 CPUs.  One estimate for the cost to achieve world championship in StarCraft II by AlphaStar comes to several million dollars once the headcount of 30 PhD-level employees as well as the computational costs are taken into account~\cite{Churchill_Buro_Kelly_2019}.\footnote{Also note that the DeepMind's achievement in StarCraft II already builds on more than a decade of open competitions run in conjunction with Game AI conferences.}
    This reference comes from work in StarCraft II that has focused on good results with bounds on what is computationally feasible (say, a single gaming PC and not a cluster of 512+ GPUs). 
\end{itemize}

One comparative disadvantage of other techniques is that over the past few years a robust set of platforms and publicly available technologies such as Tensorflow and GPU integration that have made Deep RL accessible in a way that does not apply to other techniques. This can make Deep RL much faster to implement out of the box and it is no longer necessary to implement or understand the underlying algorithms in detail.

\subsection{Evolutionary Algorithms}
Evolutionary Algorithms (EA) of various types are derivative-free \textbf{search} algorithms that start with a population of possible (often random) solutions, and make changes (mutations, cross-over) to the best of these solutions to come up with a new population of solutions; repeating this process until a sufficiently good solution is found.
A major challenge is often finding a suitable representation (genome) of the solution that is amenable to the mutation and cross-over operators.

EAs have some clear theoretical disadvantages over RL approaches; they are less sample efficient and cannot use the information from a gradient. In derivative-based RL (e.g back-propagation on Neural Networks), the policy or function being learned is updated with each individual action by gradient descent in the direction of (current) greatest benefit. In an EA the policy or function is only updated, at best, once after each full game in a random direction~\cite{Lucas_2008}.

However, EAs can be much better at exploring large search spaces precisely because of their ability to take large random jumps, and if the fitness landscape is highly irregular then gradient-descent methods can get stuck in local optima (although there are various techniques such as momentum to reduce the risk of this happening). More advanced methods such as CMA-ES or NES also incorporate a form of pseudo-derivate to make bigger jumps in more promising direction~\cite{Hansen_Ostermeier_2001, Wierstra_Schaul_Glasmachers_Sun_Peters_Schmidhuber_2014}. CMA-ES in particular has a range of software libraries across different platforms to make it accessible.
For example, using CMA-ES to evolve Deep Neural Networks has been found to be quite competitive with RL in Atari games \cite{Salimans_Ho_Chen_Sidor_Sutskever_2017}. 

Having a population of different solutions also enables niching techniques that reward a diverse set of solutions, and not just the best possible game score~\cite{preuss2015multimodal, Agapitos_Togelius_Lucas_Schmidhuber_Konstantinidis_2008, Rosin_Belew_1997}.
This is a useful technique to explore the space of viable strategies, and not just generate a single `best' one.

\subsection{Random Forests}
Random Forests are a tool used in supervised learning, especially with structured data~\cite{breiman2001random}. In Kaggle competitions they have been the most common tool used by winning competition entries, along with the related XGBoost~\cite{chen2016xgboost}, with Deep Neural Networks a close second and preferred when image data is involved~\cite{Usmani_2018}.
They consist of multiple Decision Trees, with each tree classifying the data by branching at each node on one or more of the data attributes (each node forms a binary \textit{if}\ldots \textit{then}\ldots \textit{else} condition). Aggregating the decisions of all the trees in the forest (as an ensemble method) gives much better performance than a single individual tree.
Due to the innate discontinuity at each branch in a tree, decision trees can be effective at modelling policies or functions that have similarly sharp discontinuities in them. 

Trees/Forests are not used as much in recent Game AI research, but they can be a useful tool at times. For example decision trees have been used to provide an easily interpretable policy to teach players~\cite{Silva_Togelius_Lantz_Nealen_2018}, and also as a means of learning a forward model~\cite{Dockhorn_Lucas_Volz_Bravi_Gaina_Perez-Liebana_2019}.

\subsection{Statistical Forward Planning}\label{sect:SFP}
Statistical Forward Planning (SFP) algorithms are descendants in some respects of the classical min-max search used in Chess and Checkers research in the 1980s. The combinatorial explosion of large action spaces and branching factors prevents exhaustive search to a certain depth in the game-tree, and SFP methods do not attempt this. Instead they search stochastically, and use statistics from previous iterations to make the decision about where to search next. 
At each iteration a forward model simulates the effect of possible actions on the current game state in a `what if' scenario to obtain a value for the proposed plan.

The crucial requirement is a forward model for these simulated rollouts. This is a notable difference to model-free RL methods , although model-based RL can use (or learn) a forward model to generate training data.
Even so, in \textit{vanilla} RL the learned policy that makes decisions after training is complete does not use a forward model. \textit{Vanilla} SFP methods do not do any pre-training, but use the forward model to make a decision when an action is required. They need a computational budget to use for this, and everything else being equal are slower in actual play. (See Section~\ref{sect:Hybrids} for notes on exceptions to the \textit{vanilla} case.)

Common SFP methods used in Game AI of possible relevance to wargames include:

\begin{itemize}
    \item \textbf{MCTS}. Monte Carlo Tree Search expands the game tree by one node per iteration, and stores statistics at each node on the number of times each action has been taken from that node, and the average value that resulted. These are then used to balance exploitation of the best action with exploration of actions that have been tried relatively few times. See~\cite{Browne_Powley_Whitehouse_Lucas_Cowling_Rohlfshagen_Tavener_Perez_Samothrakis_Colton_2012} for a survey.
    \item \textbf{RHEA/OEP}. Rolling Horizon Evolutionary and Online Evolutionary Planning Algorithms use EAs to generate forward plans at each iteration, based on previously successful ones~\cite{Perez_Samothrakis_Lucas_Rohlfshagen_2013, Justesen_Mahlmann_Togelius_2016}. These methods divide their budget more equally at all depths in the game-tree, which can be an advantage over the more short-term focus of MCTS in some domains.
    \item \textbf{CMAB}. Combinatorial Multi-Armed Bandits are very similar to MCTS, and have been successful in RTS games, including MicroRTS and Starcraft~\cite{Ontanon_2017}. The crucial difference is that a global action is factored into contributions from each unit under the player's control, and statistics are maintained for each unit as well as at the global level.
    
    As with any SFP, the forward model is called over a number of iterations. On each of these an action is selected for each individual unit based on past statistics to balance exploration of new actions with exploitation of previously tried good actions, exactly as in MCTS. These are then combined to give a single `global' action. At each future iteration during planning a CMAB algorithm either selects the previous best `global' action, or samples a new individual action for each unit to create a new `global' action.
    This is a good approach with a large number of units, each of which can take its own action (as in wargames). 
\end{itemize}

\subsection{Others}
There are a number of older techniques used in commercial Game AI, such as hand-written Finite State Machines, Parameterised Scripts and Behaviour Trees~\cite{yannakakis2018artificial}. These are not learned (although RL and EA methods amongst other can be used to learn the best values in parameterised scripts for example), but have advantages in ease of debugging and interpretability. 
The example of the OpenAI work on Dota2, which used hand-crafted static heuristics (or scripts) in several decision areas, shows that in an application domain these can be a useful part of a full system, even if they are not stressed in the academic literature produced.

\subsection{Hybrids}\label{sect:Hybrids}
None of these AI approaches are the answer to all games, and some will work better on some problems than others. It is better to think of them as a tools that provide options to be combined with one another as needed, with no single silver bullet that will solve each and every problem. 
Applying an AI technique to a game (or any application) first involves understanding the features and structure of the game, and tuning the appropriate tools to this structure.
Some illustrative examples of this hybridisation are listed below, and many more are covered in Section~\ref{sect:algoSummary}.

\begin{figure}
\includegraphics[width=1.0\textwidth]{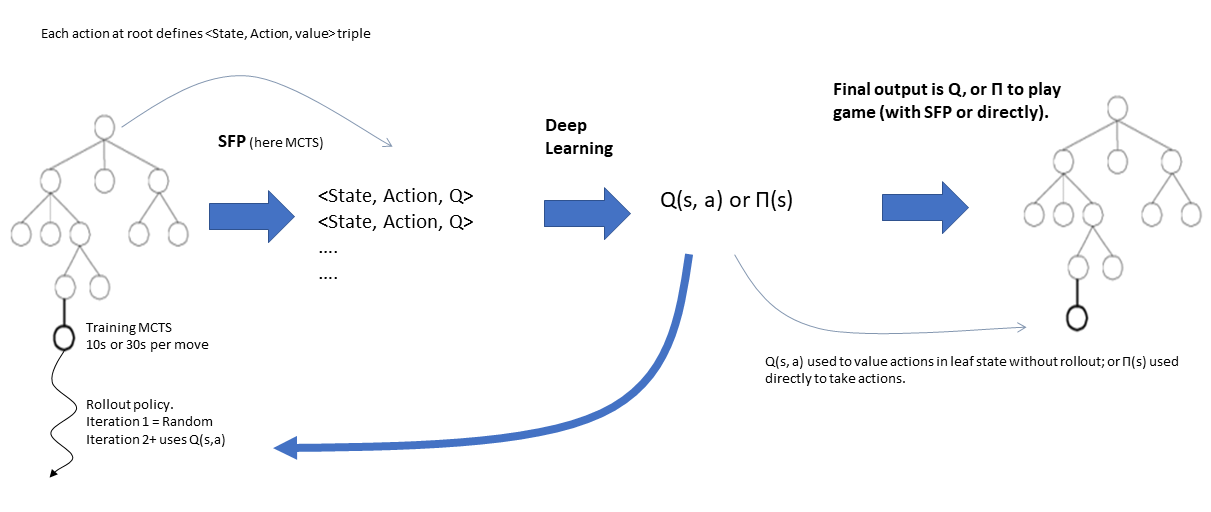}
\caption{Expert Iteration. Iterations of a Statistical Forward Planning algorithm generate data that is used to learn a policy, $\Pi(s)$, or an evaluation function $Q(s, a)$. This is then used in the next iteration of SFP to ratchet up performance.}
\label{fig:ExpertIteration}
\end{figure}

\begin{itemize}
    \item \textbf{Expert Iteration}. One particularly powerful recent hybridisation, used by AlphaGo amongst others, is between Deep RL and SFP to give `Expert Iteration'~\cite{Anthony_Tian_Barber_2017, AlphaZero_2017, Tong_Liu_Li_2019}. This iterates i) playing games using SFP to generate training data that, ii) is used by Supervised RL to learn policies or evaluation functions used in the next round of SFP games (see Figure~\ref{fig:ExpertIteration}). In this way the quality of play ratchets up, and the overall cycle can be cast a Reinforcement Learning, with the SFP algorithm used as both policy improvement and valuation operator. This is discussed further in the context of wargames in Section~\ref{sect:RelativeRS}.
    \item \cite{Marino_Moraes_Toledo_Lelis_2019} use Evolutionary Algorithms to learn an Action Abstraction that provides the Actions used in SFP (specifically, CMAB is used). This provides slightly better results than previous work by the same authors that used an expert-specified Action Abstraction directly in CMAB. This is a good example that solutions can often be built up over time, learning one aspect of a game and using rough heuristics in others. Trying to learn everything from scratch at once can make the learning task much, much more difficult by not using scaffolding that is helpful to reach a good solution before it is discarded.
    \item \cite{Stanescu_Barriga_Hess_Buro_2016} use a Convolutional Neural Network to learn the value of a state from the map in an RTS, and then use the CNN as an evaluation function in MCTS to decide which of a small number of heuristic scripts to use at that point in the game. The CNN is pre-trained on a variety of different maps so that it will generalise reasonably well to unseen scenario maps.
    \item \cite{Horn_Volz_Perez-Liebana_Preuss_2016} investigate several hybrids between MCTS and RHEA in GVGAI play. The best overall algorithm uses RHEA of the first $X$ moves and then MCTS from that point. The precise hybrid that works best varies by game, for example they find that MCTS is better at dealing with adversarial opponents in a game, but can have problem where path-finding is most important. 
    \item \cite{Lucas_Reynolds_2005} use Evolutionary Algorithms to learn Finite State Machines, and find that this is better at dealing with noisy data.
    \item A good example of constructing a system from disparate building blocks is~\cite{ha2018world}. They combine a Variational Autoencoder neural network to process a visual image, a Recurrent Neural Network as a memory component, and an evolutionary algorithm (CMA-ES) to evolve a neural network controller (policy). The first two components are trained first on game-play, and then the controller is evolved using the resultant learned model.
\end{itemize}

Algorithm are often constructed of commonly recurring components, and it can be helpful to consider how these interact in hybrids.
\begin{itemize}
    \item Policy. The most obvious component is a policy that decides what action to take in any given situation. Usually abbreviated to $\pi(s, a)$ that defines a probability of playing each action, $a$, given the current state, $s$.  Classically this is learned directly in policy-based reinforcement learning.
    
    In SFP and search-based approaches distinct policies can be used to decide which action to take in simulation using the forward model; these rollout policies tend to balance the taking the best action with exploring new actions to find a better one. 
    \item State Evaluation Function. This provides a value for a state, or for a specific action in a state. In a game this is usually the percentage change of winning, or the total score estimated if that action is taken. Usually abbreviated to $V(s)$ for a state-evaluation function, or $Q(s, a)$ for an action-state evaluation function. Classically these are learned by reinforcement learning methods, such as Deep Q-learning. (Note that a learned $Q(s, a)$ can implicitly define a policy, for example by taking the action with the highest $Q$-value; with a forward model, a $V(s)$ function can be used in the same way.)
    \item Embedding. A common technique with high-dimensional inputs is to `embed' these into a lower-dimensional latent space. This latent space is then used to make actual decisions. This is effectively what many deep neural networks, for example to convert a $10^8$-dimension visual image into a 256-dimensional embedding. This can then be used with a simple one-layer neural network, or even just a linear function, to generate a $V$, $Q$ or $\pi$ function.
\end{itemize}

\vfill
\pagebreak

\section{Algorithms for Wargames}\label{sect:algoSummary}
This section considers the distinguishing features of wargames from Section~\ref{sect:GameFeatures} in terms of the AI techniques that have been used to address these features in the recent academic literature. This covers both Deep Reinforcement Learning and other approaches. 
The objective is to provide a menu of specific techniques that are useful for wargames in particular, and provide some references to these. It feeds into the recommendations of the report, but is also intended to be a useful reference resource in its own right.

\begin{table}[h!]
\begin{tabular}{|p{3.5cm}|c|p{8cm}|}
\hline
\textbf{Feature} & \textbf{Section} & \textbf{Impact, key techniques and references}                                         \\ \hline
Forward Model & \ref{sect:ForwardModel}  & Statistical Forward Planning (MCTS, CMAB, RHEA etc.)~\cite{Browne_Powley_Whitehouse_Lucas_Cowling_Rohlfshagen_Tavener_Perez_Samothrakis_Colton_2012, Ontanon_2017, Perez_Samothrakis_Lucas_Rohlfshagen_2013}. Expert Iteration~\cite{Anthony_Tian_Barber_2017, AlphaZero_2017}. Semantic representations are usable instead of pixel-based ones. \\ \hline
Unit-based Actions with large State Space & \ref{sect:unitTechniques} & Factorisation of the problem into individual unit decisions. Hierarchical methods. See text for references. \\ \hline
Large Action Space  & \ref{sect:ActionSpace} & Action Abstraction, High-level script-based action definitions~\cite{Churchill_Buro_2013, Neufeld_Mostaghim_Perez-Liebana_2019, Moraes_Marino_Lelis_Nascimento_2018}.                                             \\ \hline
Information Flow and Unit Autonomy &  \ref{sect:InfoFlow2}  & DecPOMDPs~\cite{oliehoek2016concise, Melo_Spaan_Witwicki_2012}, Coevolution~\cite{chades2002heuristic}, Unit-factorisation techniques (of Observation as well as Action space). Graph Neural Networks~\cite{zhou2018graph}\\ \hline
Imperfect Information (Fog of War)                  & \ref{sect:POMDP}            & Particle Filters~\cite{Doucet_Johansen_2009, Silver_Veness_2010}, Monte Carlo methods~\cite{Browne_Powley_Whitehouse_Lucas_Cowling_Rohlfshagen_Tavener_Perez_Samothrakis_Colton_2012, Cowling_Powley_Whitehouse_2012}, POMDP Planning~\cite{Amato_Oliehoek_2015, Ross_Pineau_Chaib-draa_Kreitmann_2011, Silver_Veness_2010}.  \\ \hline
Adversarial Opponent  \& Player Asymmetry &  \ref{sect:AdvOpponent}, \ref{sect:PlayerAsymmetry}  &   Hall of Fame and Tournaments~\cite{Cliff_Miller_1995, Moriarty_Mikkulainen_1996, Rosin_Belew_1997}. Asynchronous Deep Learning~\cite{mnih2016asynchronous}. Co-evolution~\cite{Rosin_Belew_1997}. Opponent Modelling~\cite{Albrecht_Stone_2018}. \\ \hline
Relative Reward Sparsity               & \ref{sect:RelativeRS}     & Reward Shaping, Expert Iteration~\cite{Anthony_Tian_Barber_2017, AlphaZero_2017}, Intrinsic Motivation, Curriculum methods~\cite{Bengio_Louradour_Collobert_Weston_2009, sukhbaatar2017intrinsic, Justesen_Risi_2018}\\ \hline
Scenario Variability (and RL overfitting)              &  \ref{sect:ScenarioVar}, \ref{sect:overfit} &  Statistical Forward Planning. Meta-learning and Curriculum learning~\cite{brant2017minimal, Bengio_Louradour_Collobert_Weston_2009}                                                                          \\ \hline
Objective Variability \& Strategy Exploration  &  \ref{sect:Exploration}  & Multi-Objective optimisation~\cite{Zhou_Qu_Li_Zhao_Suganthan_Zhang_2011}. Niching and Diversity Maintenance~\cite{preuss2015multimodal, Mouret_Clune_2015, Rosin_Belew_1997}.                                             \\ \hline
Map-based &  \ref{sect:maps}  & CNNs, Conflict-based search~\cite{boyarski2015icbs}.\\ \hline
Human-like play / doctrinal control &  \ref{sect:imitation} & Imitation Learning~\cite{Ross_Pineau_Chaib-draa_Kreitmann_2011}, Inverse RL~\cite{Ng_Russell_2000}, Preference Based RL~\cite{wirth2017survey}\\ \hline
\end{tabular}
\caption{Summary of the key distinguishing features of wargames, and which impact the use of AI techniques.}
\label{table:KeyFeatures}
\end{table}
Table~\ref{table:KeyFeatures} lists the main features of wargames that most impact the use of AI techniques. It is a summary of the distinctive elements of wargames teased out of the more exhaustive review of Section~\ref{sect:GameFeatures}. 
The remainder of this section goes into these areas in more detail and reviews AI techniques that address these features, and which are therefore likely to be of most use in wargames.

\subsection{Forward Model}\label{sect:ForwardModel}
The forward model of a game takes the current state and the current set of actions, and transitions to the next state.  Every computer game has such a model; without it the game would not function.  However, in order to use the model in statistical forward planning and model-based reinforcement learning algorithms we need it to be both fast and easily copied.  The model needs to be easily copied so that hypothetical actions can be applied to a copy of the game rather than the live game; this can be technically demanding due to the need to be very careful about which parts of state need to be deep-copied and is less bug-prone with heavy use of immutable data types.  Speed is important for SFP algorithms in order to run many simulations for each decision to be made in the live game.  For model-based RL, speed is less important but also useful due to the high sample complexity of these methods.
With a copyable forward model a number of statistical planning-based algorithms are usable, such as Monte Carlo Tree Search (MCTS)~\cite{Browne_Powley_Whitehouse_Lucas_Cowling_Rohlfshagen_Tavener_Perez_Samothrakis_Colton_2012}, Combinatorial Multi-Armed Bandits (CMAB)~\cite{Ontanon_2017}, and Rolling Horizon Evolutionary Algorithms (RHEA)~\cite{Perez_Samothrakis_Lucas_Rohlfshagen_2013}. These are reviewed in Section~\ref{sect:SFP}, and are especially useful if there is room for `thinking' time between each move, anything from 40 milliseconds to a couple of minutes.

A fast non-copyable forward model is still useful for a subset of AI techniques, most notably Deep Reinforcement Learning. In this case the game can only be run end-to-end from a configured starting point, and planning algorithms from a mid-point cannot be used. But each of these end-to-end runs can be treated as a single iteration to generate data for RL algorithms. For example OpenAI's state of the art Dota2 player is trained using Dota2 games that run at approximately half normal speed (for local optimisation reasons) with no copyable forward model~\cite{berner2019dota}.

If we do have a copyable fast forward model then Expert Iteration is a technique that DeepMind used in AlphaZero to great effect, and which combines Deep RL with MCTS. This has also been helpful in other games such as Hearthstone~\cite{Swiechowski_Tajmajer_Janusz_2018} and Hanabi~\cite{Goodman_2019}, and is discussed in Section~\ref{sect:OtherAlgos}.

Linked to the existence of a forward model is that in any game we have access to a semantic representation of the game state (unit positions, current capabilities, sighted enemies, terrain details etc.). In this case it is generally preferable to use these as input features for any technique, including Deep RL, and not pixel-based screen shots. This avoids the overhead of learning an implicit semantic representation from pixels and means that the AI technique can focus on learning an optimal plan directly. The prevalence of pixel-based inputs in the academic literature even when a forward model is available is due to the underlying goal of Artificial General Intelligence (AGI) that learns from raw sensory input. This is not relevant to wargames. 
Pixel-input is also much used in the literature due to the focus on self-driving technology, for which there is no forward model and raw sensory input is all that is available. Again, this is not relevant to wargames.

\subsection{Unit-based techniques}\label{sect:unitTechniques}
There are two distinct aspects of the use of units in wargames that lead to slightly different groups of techniques. These are:
\begin{enumerate}
    \item \textbf{Compositional/Factorisation} techniques. These consider that rather than trying to come up with a single policy or plan across all available units it is quicker to come up with a plan for each unit individually. A set of individual unit plans/policies will be less effective than a global plan, but is much less computationally demanding and may be more akin to the way that unit operate in `reality'. 
    \begin{itemize}
    \item{Combinatorial Multi-armed bandits (CMAB). Where multiple units can be given simultaneous orders each underlying unit action can be selected independently (a unit-based `arm'), and these amalgamated into a global `arm'~\cite{Ontanon_2017, Barriga_Stanescu_Buro_2017}. The global level allows for dependencies between units to be learned. }
    \item{Multi-Agent Reinforcement Learning (MARL) that learns a policy for each unit independently. The idea again is to benefit from the fact that the state space at least partially factorises into components for each unit, and learning these factorisations will be faster than the joint policy~\cite{Tan_1993}.}
    \item{Decentralised POMDPs~\cite{oliehoek2012decentralized, Amato_Oliehoek_2015}. This is an alternative way of phrasing the problem using a unit-based factorisation, however the literature generally focuses on problem-solving in a non-adversarial environment such as Fire Brigade units co-operating to extinguish a fire~\cite{Paquet_Bernier_Chaib-draa_2004}. Units are generally, but not always, homogeneous, with the benefit coming from learning an `easier' policy for a generic individual unit instead of a more complicated global policy.}
    \item Search a game-tree for each unit individually, and then amalgamate these results~\cite{Lisy_Bosansky_Vaculin_Pechoucek_2010}. A similar reduction of the combinatorial problems of many units is to group units together into smaller sub-groups, for example based on proximity or unit type, and only regard interactions in planning actions within each sub-group.
\end{itemize}
    \item \textbf{Hierarchical} techniques. These consider decision-making at multiple levels, and is a natural fit for many aspects of wargaming. A higher-level strategic policy makes decisions about broad goals and transmits these goals to the lower-level units, which seek the best local way to implement them. This hierarchy can theoretically have many levels, but two is usually enough in the literature.
\begin{itemize}
    \item Hierarchical approaches that set a high-level goal (e.g. `Attack', or `Defend Position'), then grouping of units to a sub-goal, and at the lowest level those specified units planning/learning to achieve their sub-goal~\cite{Stanescu_Barriga_Buro_2014, Churchill_Buro_2015}. Different algorithms can be used at each level in the hierarchy, varying between heuristic (expert) rules, Deep RL, tree-based search and others. This decomposition of the overall task reduces the inherent combinatorial explosion, with lower levels using a reward function to achieve a goal specified by the previous level. 
    \item A similar approach can be used to decide which evaluation function to use in MCTS (or any search algorithm), for example by using Hierarchical Task Networks as the higher planning level to select which evaluation function to use in MCTS at the lower execution level~\cite{Neufeld_Mostaghim_Perez-Liebana_2019}. 
    \item In the work of \cite{Ontanon_Buro_2015} in an RTS game, 19 high-level `tasks' (e.g. harvest resources, or rush player) and 49 low-level `methods' are used in hierarchical planning. By using A* pathfinding and `methods' as the lowest level in planning rather than primitive actions (such as `move north one hex') this technique leverages domain knowledge and performs well compared to MCTS or other methods. 
    \item FeUdal Networks use Deep Reinforcement Learning with a `Manager' network setting a high-level goal, and then `Worker' network(s) attempting to fulfil the high-level goal~\cite{Vezhnevets_Osindero_Schaul_Heess_Jaderberg_Silver_Kavukcuoglu_2017, ahilan2019feudal}. This could be adapted to a wargame environment with one Worker network per unit, or group of units.
\end{itemize}
\end{enumerate}

\subsection{Information Flow and Unit Autonomy}\label{sect:InfoFlow2}
There is relatively little academic work on restricted Information Flow within Game AI, outside of more abstract games with deliberately constrained information channels, such as Hanabi or Bridge. In these games Deep Learning approaches have been used to learn conventions from scratch~\cite{Foerster_Song_Hughes_Burch_Dunning_Whiteson_Botvinick_Bowling_2018, Yeh_Hsieh_Lin_2018}. 

More relevant to wargames is work in Dec-POMDPs (Decentralized Partially Observable Markov Decision Processes) that have multiple autonomous agents on the same team working towards a single team-level goal~\cite{oliehoek2016concise, Paquet_Bernier_Chaib-draa_2004, Melo_Spaan_Witwicki_2012}. These were mentioned in Section~\ref{sect:unitTechniques} as appropriate to reducing the complexity of managing multiple different units by factorising a plan into semi-independent plans for each unit, and then amalgamating these to give a global plan.
Within this framework there are several options that can be used to additionally model unit autonomy and information flow:
\begin{itemize}
    \item \textbf{Global Information}. The default option is to use global information for each individual decision/plan. This provides benefit in reducing the computation of the overall plan, but assumes that each individual unit has the same information as the overall commander.
    \item \textbf{Local Information}. Each unit makes its decision using only locally available information. This models local autonomy more directly, is realistic in the case of a unit being cut-off from communication networks, and is computationally cheaper.
    \item \textbf{Information Cost}. Communicating information can be treated as an action in its own right, and the commander can choose to pass on some global information to units that they can then incorporate in local decisions. This can also work in the opposite direction. There is little work in this area, and the focus is on side-ways communication between the individual agents without any central command unit~\cite{Melo_Spaan_Witwicki_2012, wu2011online}.
\end{itemize}

Consider a concrete example of how the framework might be adapted in a wargame environment to provide low-level control of units once given a target by the commander (either human or computer). If guided by a utility function that applies suitable weights to secondary goals such as self-preservation and remaining unobserved as well reaching the target, then instead of following a centrally-plotted path the units could make their own decisions at each point based on their local information only, adapting this as they become aware of enemy (or friendly) units. 

There are no restrictions on the algorithms used for individual agent policies in a Dec-POMDP setting, and RL or search-based planning methods are both common.
Co-evolution is also used~\cite{chades2002heuristic}, and this can be especially relevant with an absence of formal communication and heterogeneous units, where each unit type needs to adapt to the others.

Graph neural networks (GNN) are a new emerging area of research in Deep RL that may be a useful tool for modelling information flow in wargames~\cite{zhou2018graph}. The basic idea behind a GNN is to model the dependencies of graphs via learned message passing between the nodes of the graph. The same neural network is used for all nodes in the graph to learn to integrate information from incoming nodes and to iteratively come up with a solution.
In the case of wargaming, nodes in the GNN could represent units, while the connections between them reflect communication channels. A GNN could be trained to integrate information in a decentralised way and learn by itself to resolve communication issues. GNNs are discussed in a little more detail in Section~\ref{sect:GNN}.

\subsection{Imperfect Information}\label{sect:POMDP}
More realistic algorithms that work in large-scale POMDPs usually sample from some estimate of the unknown State Space: 
\begin{itemize}
    \item Perfect Information methods can be used by sampling possible states, then solving each sample, and then taking the `average' of the best actions. These approaches have theoretical flaws, but can work well in practise in some domains. To understand the detail of the flaws (`strategy fusion' and 'non-locality'), see~\cite{Frank_Basin_1998}.
    \item Monte Carlo search methods sample from possible hidden states and take an expectation over these. For example Information-Set MCTS samples one possible state of the world on each iteration~\cite{Cowling_Powley_Whitehouse_2012}, and Imperfect Information Monte Carlo Search (IIMC) does the same but with recursive opponent models that factor in what the opponent can observe to allow bluffing~\cite{Furtak_Buro_2013}.
    \item A distribution over hidden information (or `belief states') can be maintained with a  particle filter and updated as the game unfolds to complement these methods~\cite{Silver_Veness_2010}. This approximates a full Bayesian update, which is usually intractable in a non-trivial game environment.
    \item Counterfactual Regret Minimisation (CFR) has worked exceptionally well in dealing with hidden information and player bluffing in up to 6-player Texas Hold'em Poker~\cite{Zinkevich_Johanson_Bowling_Piccione_2008, Lanctot_Lisy_Bowling_2014, Brown_Sandholm_2019}. However this required pre-calculation of a whole-game Nash Equilibrium strategy, which took 8 days on 64 cores for 6-players. Some work has compared MCTS to Monte Carlo versions of CFR, and found that while CFR can give better results, MCTS performs better for more limited computational budgets~\cite{Ponsen_2011}.
    \item Deep Reinforcement Learning can be used directly on the Observation Space, relying on implicit models of the hidden state to be built. This applies in Dota2 and also in other approaches to Poker~\cite{Deepstack_2017, berner2019dota}.
    \item Smoothed versions of both MCTS and Deep Learning play a mixture of a best response policy with an averaged fictitious play can help convergence in imperfect information environments~\cite{Heinrich_Silver_2015, Heinrich_Silver_2016}.
\end{itemize}

\subsection{Adversarial Opponents}\label{sect:AdvOpponent}

The presence of adaptive, adversarial agents complicates any learning approach and introduces issues around the `Red Queen' effect, in which improving play will not necessarily improve the outcome if the opponent is also learning.
This is a well-documented problem in co-evolution, with non-progressive cycles. Consider a simple case of Rock-Paper-Scissors. If the opponent has a tendency to play Rock, then we will learn to play Paper, which will prompt the opponent to learn to play Scissors\ldots and we can keep cycling \textit{ad infinitum}. 
It is important to use a variety of training opponents to avoid over-adapting the current policy to the current opponent. This is sometimes called a `Hall of Fame' and can be constructed from policies from earlier stages of the learning process~\cite{Cliff_Miller_1995, Moriarty_Mikkulainen_1996, Rosin_Belew_1997}. This ensures that we do not forget how to deal with particular strategy once we have defeated it once. In the Rock-Paper-Scissors example this would learn a strategy that can be effective against an opponent that plays any of Rock, Paper or Scissors, and converges to the Nash Equilibrium strategy.

Approaches of this sort are used in AlphaStar, with multiple agents learning in parallel and playing each other in a Tournament to bootstrap strategies against each other without over-fitting to a single opponent (see Section~\ref{sec:continual_learning}). It is notable that master level Go in AlphaZero did not require this opponent variety and only used self-play against the current policy. However, AlphaZero (unlike the pure RL of AlphaStar) used MCTS planning on top of RL to select each move, which would be able to detect longer-term problems in a highly recommended move at the top level of the tree.

AlphaZero did use distributed asynchronous learning, as did OpenAI's Dota2 work. This involves running many thousands of games simultaneously, each using a local copy of the current learned policy and providing updates to a central hub. This hub learns the policy using the traces of all games and periodically sends out updates. This helps avoid overfitting of the policy by including information from many different games at different stages in each update~\cite{mnih2016asynchronous, berner2019dota}.

Another tool to consider in an adversarial environment where both sides are learning is the `Win or learn fast' principle~\cite{Bowling_Veloso_2002}. Roughly speaking, this ramps up the learning rate when we are losing training games compared to when we are winning. This gives better theoretical convergence guarantees in many cases.

Opponent Modelling is the general idea that we think about what our opponent will do when planning our moves~\cite{Albrecht_Stone_2018}. In the context of a 2-player primarily zero-sum wargame the best approach is to learn the opponent's strategy simultaneously (with Deep RL or co-evolution), or use the same planning technique (e.g. MCTS). More sophisticated `theory of mind' approaches are unnecessary, and are primarily of benefit in general-sum games where there is an incentive for players to co-operate in at least some parts of the game.

\subsection{Relative Reward Sparsity}\label{sect:RelativeRS}

It is a triumph of RL (and especially DeepRL) that it is able to cope so well with such sparse rewards, and learn policy and value networks able to play classic games to a superhuman standard (see AlphaZero~\cite{AlphaZero_2017} and MuZero~\cite{schrittwieser2019mastering}).  The algorithms rely on the fact that although the rewards
are sparse, there are patterns in the game state that can accurately predict
the value, and these patterns can be learned. It is essential that during learning, the learner
gets to experience a range of outcomes; for example, if every playout ends in a loss, there is no learning signal.
  
In RL if the inherent reward signal is too sparse then \emph{Reward Shaping}
may be used to provide additional incentives and guide a learning
agent towards successful behaviours.  Related concepts are curriculum
learning, where the agents learns tasks of increasing complexity
on the path towards the full scale of problem, and intrinsic motivation, where the agent actively seeks out novel game states it has not seen before~\cite{bengio2009curriculum, sukhbaatar2017intrinsic, Justesen_Risi_2018}.

$\epsilon$-greedy or softmax approaches often do very poorly with sparse rewards because they explore independently at each action/time-step. Instead we want to use an exploration policy that is `consistent' over time, making similar deviations when exploring, for example by biasing moves that go up-hill, or that attack at worse odds than normal.
There is a rich literature on how we can encourage consistent exploration in RL, including evolution of the evaluation function or policy and intrinsic motivation to seek out novel states~\cite{Bellemare_Srinivasan_Ostrovski_Schaul_Saxton_Munos_2016, Ostrovski_Bellemare_Oord_Munos_2017, Salimans_Ho_Chen_Sidor_Sutskever_2017, Osband_Aslanides_Cassirer_2018, Osband_Blundell_Pritzel_VanRoy_2016}. 

Expert Iteration learns an interim reward signal that can be used in short-term planning towards a long-term goal that is beyond the planing horizon. This is a technique used in AlphaZero (amongst others) to learn a game evaluation function from RL, and then use this in MCTS search~\cite{Anthony_Tian_Barber_2017, Swiechowski_Tajmajer_Janusz_2018}. Over several iterations of this meta-process the local evaluation function is bootstrapped up to learn what local, short-term features are conducive to global, long-term success.
Model Predictive Control (MPC) is another option if we have a forward model. This uses a planning horizon of $H$ to plan an optimal sequence of moves, and then re-plans after taking the first of these and seeing the effect in the environment~\cite{Lowrey_Rajeswaran_Kakade_Todorov_Mordatch_2018}. It is possible to learn an evaluation function (via RL for example) that is then used to optimally plan up to $H$ steps ahead using MPC.
Learning an interim evaluation function can also provide useful information on what the agent is seeking to maximise in the short-term as useful to achieve a long-term goal, as the weights it uses may be quite different to those a human expert would design in.
This is not always needed. For example the world-championship beating Dota2 work from OpenAI used a hand-designed short-term reward signal based on the team's domain knowledge. Where this domain knowledge exists, this is usually a better thing to try first, as it can give the desired level of play with less complexity.

\subsection{Strategy Exploration}\label{sect:Exploration}
In some use-cases the purpose of a wargame is not to find the single best course of action, but to explore a range of possible outcomes. This is especially true in the case of Concept or Force Development wargames.
A number of techniques can be useful here to ensure that a diverse range of policies is found:

\begin{itemize}
    \item \textbf{Multi-objective optimisation}. This starts from the premise that there may be multiple different objectives, and a policy is needed that can adapt at run-time to the actual objective. It hence seeks a range of different policies that are `optimal' in different ways~\cite{Deb_Agrawal_Pratap_Meyarivan_2000, Perez-Liebana_Mostaghim_Lucas_2016}. 
    Given a set of different objectives, for example to a) minimise casualties, b) minimise fuel usage, c) maximise the enemy forces degraded, a multi-objective approach returns a set of policies on the `Pareto-front' of these dimensions. 
    A policy on the Pareto-front is one which is better than all other policies for at least one linear combination of the different objectives; for example with 80\% weighting on minimising casualties, 20\% on fuel usage and 0\% on degrading the enemy. 
    These can also be set as secondary targets to the main goal of the wargame to give policies which achieve the core victory objective with the best result on this secondary target. 
    When the actual objective of a game is known (usually at game start, but this could change at some mid-point in line with Objective Variability discussed in Section~\ref{sect:ObjectiveVar}), then a relevant policy to be used can be selected from the trained library. See~\cite{mossalam2016multi} for a Deep Learning multi-objective approach.
    \item \textbf{Niching}. In Evolutionary Algorithms that maintain a population of policies, an additional bonus can be specified to ensure that this population remains diverse and does not converge too quickly on a single solution. This helps ensure that more of the possible policy space is explored. Classic options here can include a reward based on genome-distance, if this can be specified, but this may not automatically correspond to phenotypic distance. An alternative approach is to match the population against a diverse set of opponents (most commonly from a `Hall of Fame' from past generations), and divide the reward for beating each across all those population members that are able to. This rewards a policy that can defeat one or two opponents the others cannot, even if it has a low overall score~\cite{Rosin_Belew_1997}. For a more detailed review see~\cite{preuss2015multimodal}.
    \item \textbf{MAP-Elites}. Similar to Multi-objective optimisation, MAP-Elites is a recent algorithm that seeks to find diverse different solutions to the same problem~\cite{Mouret_Clune_2015}. It requires the specification of one or more dimensions of `phenotype', which for example could be `movement', `fuel usage' or `casualties suffered'. These are combined in a grid, and an optimal policy is sought for each cell (e.g. high movement, low fuel usage and low casualties).
    \item \textbf{Nash Averaging}. This can be used in combination with any of the above approaches, as well as in its own right. The common intuition is that a set of different policies may not be very diverse in practise. Nash Averaging can use game theoretic analysis on the results of tournaments between the policies (or their performance on a range of different scenarios) to extract the small sub-set that between them encapsulate the full diversity of the population~\cite{Balduzzi_Tuyls_Perolat_Graepel_2018}.  Nash Averaging has the benefit of discounting duplicate opponents when assessing the strength of each player, and hence prevents a weak player appearing strong by being especially good at beating a cluster of self-similar weak players.
\end{itemize}

\subsection{Map-based}\label{sect:maps}
One feature of many wargames, especially land-based ones, is that they are played out on a map of varying terrain features. There are a few approaches that are natural tools to use in this case, which we highlight here:
\begin{itemize}
    \item \textbf{ Convolutional Neural Networks (CNNs)}. While we strongly recommend against learning to play wargames from raw pixel visual inputs, CNNs are still a useful tool for processing a symbolic version of a map. This is for example what a Go or Chess board really are. CNNs are used for example in~\cite{Barriga_Stanescu_Buro_2017} to process a symbolic representation of an RTS map to pick one of four scripts to use for a given unit.
    \item \textbf{Multi-unit path-finding}. A* is the classic algorithm in this case, but copes less well when multiple units are involved. In this case Conflict-Based Search is the current state-of-the-art~\cite{sharon2015conflict, boyarski2015icbs}.
\end{itemize}

\subsection{Human-like control}\label{sect:imitation}

In some wargame use-cases (see Section~\ref{sect:wargameTypes}) there is a requirement for an AI to act in a `human-like' way, which may also involve following standard military doctrine for the force in question.
Deep RL that has allowed agents to beat the world champions in some games, is less relevant for these use-cases, in which policies should not perform arbitrary policies but policies that are simultaneously high-performing and human-like. 
To encourage more human-like play, data from humans can be incorporated at different points in the training loop. 

\textbf{Imitation learning:} One typical approach is to collect human playtraces and then train a policy network in a supervised way to reduce the error between the predicted action for a certain state, and the action that a human performed. This approach was part of the original AlphaGo, in which a policy was trained to predict the most likely actions human would perform using databases from master-level tournaments~\cite{AlphaGo_2016}. This policy was then used to guide the MCTS tree-search by seeding initial values for each leaf node. 
In this example the method was used to get a better policy, not one that was specifically `human-like', and in later work (AlphaZero), the authors discard this approach~\cite{AlphaZero_2017}. 
Other examples have aimed at `human-like' behaviour via this approach, for example learning a player-specific driving style in a racing game~\cite{Munoz_Gutierrez_Sanchis_2013}.

Beyond very simple cases `naive' imitation learning in this fashion can be fragile and lead to very poor performance when the agent moves even a small distance from the training data. To mitigate this algorithms such as \textsc{DAgger} can be used to generate new expert-like training data in situations where this deviation is most likely~\cite{Ross_Pineau_Chaib-draa_Kreitmann_2011}. This requires a policy that can give a reasonable 'expert' move in any given situation.
In a wargame context human experts could indicate what the `correct' move is after the AI has taken one in a few trial games, and this used to augment the training data.

\textbf{Reward shaping:} Another approach, which was deployed in AlphaStar, is to incorporate the distance to human playtraces into the reward function used for RL. For example, in AlphaStar, bots were rewarded for winning while at the same time producing strategies reminiscent of human strategies. The benefit of this approach is that it directly allows the agent to optimize for both metrics, without needing a dedicated purely supervised pre-training phase. 
This approach requires a suitable metric to be defined between human play and the strategy followed, and this often requires expert domain knowledge to specify what classic human strategies look like in game terms.

\textbf{Inverse Reinforcement Learning:} This approach `inverts' the classic RL problem by seeking to learn the reward function an agent is using based on their behaviour~\cite{Ng_Russell_2000}. It is useful for tasks in which a reward function might be hard to define. 
Human playtraces are used to estimate the implicit reward function, and then this reward function is used in classic RL to learn a policy. This avoids some of the fragility issues with naive imitation learning, but is computationally more demanding. 
This approach has been used, for example, to construct a human-like AI in the game of Super Mario~\cite{Lee_Luo_Zambetta_Li_2014}.

\textbf{Preference Based Reinforcement Learning: } A reward function can also be learned from human preferences~\cite{wirth2017survey}. In the approach by~\cite{christiano2017deep}, humans are presented with short video clips of agents performing   Atari or MuJoCo robot control tasks, starting from random policies. The user selects which of the two policies they prefer; by repeatedly asking a human for the preferences, a model of the user's preferences is learned through supervised learning. Once this model is learned, RL can be driven by this learned model. This approach was successful applied to learning policies for different simulated robots. For our wargaming purposes, we imagine units could be trained through RL via self-play and a first estimate of a reward function. Every once in a while a user is asked to compare selections of agent policies, which should allow to refine the initial reward function to produce more and more human-like and realistic strategies over time. 

\vfill
\pagebreak

\section{Platforms}\label{sect:Platforms}

In this section we review the main Game AI platforms.  In this context a platform is a system that combines the following:

\begin{itemize}
\item An extensible set of games on which to test agents.  To be relevant to war games, support for multi-player games is essential.
\item An extensible set of agents to play the games.
\item One or more ways to evaluate the agents on the games.  Multi-player games have the added complexity of needing to pit each agent against many other agents on a range of games.
\item A well defined set of interfaces to define how the agents interact with the games, and how the evaluation framework can instantiate the games and the agents and run them.
\end{itemize}

Beyond this, there is an important point which all of the current platforms offer little or no support for:

\begin{itemize}
    \item Parameter tuning
\end{itemize}

All of the game AI algorithms, and all of the games have parameters that can be tuned\footnote{When parameters control how other parameters are set, they are often called hyper-parameters.  For example, the learning rate used to train a neural network controls is often called a hyper-parameter, as it controls how the weights (parameters of the network) are set.}.
For the Game AI algorithms, the purpose of parameter tuning is to generate the desired level of performance, as far as is possible.  For the games, tuning can serve many purposes, including optimising the balance of a game, or to find the circumstances under which particular strategies succeed or fail.

Given the importance of parameter tuning, we recommend that special attention be paid to this, and that a standard interface be developed to interface between the optimisation algorithms (e.g. NTBEA \cite{NTBEA-Game-Tuning} \cite{NTBEA-efficient-opt}) and the entity being tuned (whether agent or wargame).  The first step is to do this for single objective optimisation problems with a fixed objective.  Beyond this is multi-objective optimisation (which often provides a better fit to the underlying problem), and problems where the measure of success is not fixed, but depends on the other agents or games under test, which is naturally the case when evaluating agents on multi-player games.  This measure of success is standard on co-evolutionary learning, and also relevant is the concept of Nash Averaging used in AlphaStar\cite{Balduzzi_Tuyls_Perolat_Graepel_2018}.

We now describe all the platforms we consider to be most relevant to wargame AI.  As we shall see, the ideal platform for wargame AI does not yet exist, though many of the ideas needed are already present in one or more of the current platforms.

\subsection{Open AI Gym}
\url{https://gym.openai.com/}

This is a widely used framework within reinforcement learning and includes a wide variety of environments, such as the Atari Learning Environment (ALE) and MuJoCo tasks.  However, there are significant limitations: support for planning algorithms is limited (e.g. the copy() method, essential to statistical forward planning algorithms) works on the ALE tasks, but not on others.  The main limitation which renders it unsuitable for wargaming is the lack of support for multi-player games.

\subsection{OpenSpiel}

\url{https://github.com/deepmind/open_spiel}

\begin{itemize}
\item Core implementation language: C++, with wrappers for Python (and possibly other languages)

\end{itemize}

OpenSpiel is led by Marc Lanctot from DeepMind.  The main focus of OpenSpiel is on classic board and card games, though it can potentially be used by any game that meets the standard interface and has an SFP-ready game state (a forward model that can be easily copied and run fast).  The standard interface
assumes that actions in a game can be enumerated.  At each agent decision point, the set of available actions is passed to the agent and the agent 
then selects one of them.  This is 
a standard interface that works well for classic
games with small numbers (e.g. up to hundreds) of possible moves in each game state, but not for all games.  

For example, in Flashpoint Campaigns, move actions can contain up to three waypoints en route to the destination, each of which can be on any hex in the map.  Hence, if the map has 100 hexes, then there would be $10^8$ possible movement paths for each unit, which would be infeasible to enumerate.

OpenSpiel has an elegant way of representing chance events, which are a central feature of games such
as Backgammon and Poker.  Chance is represented as another player in the game, end each chance event
is modelled as a move made by the chance player.
Each state in the game has a one to one mapping from the state to the move history, which is made possible by the inclusion of the chance moves in the move history.

One of the main attractions of OpenSpiel is the significant number of implementations of recent Game AI algorithms, such as Monte Carlo Tree Search and a range of DRL agents such as DQN and A3C.  OpenSpiel also comes with a wide range of classic games, which are parameterised to enable different versions to be run.

\subsection{Polygames}

\url{https://github.com/facebookincubator/Polygames}

\begin{itemize}
\item Core implementation language: C++, with wrappers for Python (and possibly other languages)
\end{itemize}

Polygames is similar in scope to OpenSpiel, with the main focus on 2-player classic board games such as Go and Othello.  Compared to OpenSpiel, it has fewer games and fewer AI agents.  However, Polygames does provide a standard way to interface convolutional neural nets to board game environments.
As with OpenSpiel, games in Polygames can be run with different parameters (such as board size).   Although the challenge of learning from pixels (and more generally from visual input such as board-tiles) is less relevant to wargame AI, it may still be helpful to allow agents to learn from a spatial arrangement of map-tiles.

\subsection{ELF: Extensive, Lightweight and Flexible platform for game research}

\url{https://github.com/facebookresearch/ELF}

\begin{itemize}
\item Core implementation language: C++, with wrappers for Python (and possibly other languages)
\end{itemize}

ELF comes with three sample enviromments: an ALE wrapper, Go, and most relevant to wargames, a mini RTS.  The mini-RTS game is similar in nature to microRTS.  ELF offers a reasonably fast and well designed implementation in C++, but does not offer any particular advantages for wargames.

\subsection{General Video Game AI (GVGAI)}
\url{http://gvgai.net}

GVGAI \cite{GVGAI-Multi} is a multi-faceted game AI framework 
that enables experimentation with many
aspects of video games, each defined
by its own competition track:

\begin{itemize}
    \item The Single-player planning track. Agents use the forward model to play a wide range of games and puzzles.  Leading agents include a range of tree search methods as well as rolling-horizon evolution.  This track has had the most interest.
    \item Single-player learning track.  This is most similar to the ALE framework, and agents are given a short learning period on some training levels before attempting to play unseen test levels.  The short learning time and limited advance information made this (in retrospect) too challenging and led to a limited number of entries.  Agents can learn from pixels or an object-based game state observation.
    \item Two-player planning track: agents have access to the forward model, but have to cope with an unknown agent on an unknown game.
    \item Level-generation track: the aim is to generate interesting (or at least playable) novel levels for unseen games: very challenging.
    \item Rule-generation track: generating new game rules for novel games; again very challenging.
\end{itemize}

\begin{itemize}
\item Original implementation in Python (Schaul), more recent implementation in Java (much faster and more competition/evaluation tracks (2-player, learning track, level generation track).  Java version also has wrappers for Open AI Gym
\end{itemize}

The learning track of GVGAI is similar in nature
to ALE.  ALE has the advantage of using 
commercial games (albeit from a bygone era)
that are easy to relate to, though due to the
finite set, limited number of levels for each 
game and deterministic nature has been susceptible to significant over-fitting 
\cite{OpenAIOverfitting}.
Although less widely used than ALE,
GVGAI does have some distinct advantages
that are relevant to Wargame AI:

\begin{itemize}
\item An extensible set of games, game levels
and game variations which are relatively easy to author in VGDL
\item Support for 2-player games
\item Stochastic games
\end{itemize}

A particular feature of the GVGAI framework is the \textit{Video Game Description Language} (VGDL).  VGDL is a Domain Specific Language (DSL) designed to make it easy to express the rules of typical arcade-oriented 2D video games.  Although VGDL was designed with this in mind, the overall approach of having a high-level DSL is very relevant for war game AI, and one of our recommendations is to consider developing a language to describe war game scenarios, including maps, assets, objectives and observers (where an observer specifies how and when to convert game states in observations that can be fed to the agents).
\footnote{With languages such as Kotlin implementing DSLs is straightforward, and enables edit-time type-checking which greatly assists in
the content authoring process:  \url{https://kotlinlang.org/docs/reference/type-safe-builders.html }}

\subsection{Codingame}
\url{https://www.codingame.com/start}

Codingame is a bit of an outlier here in this section as it does not meet the criterion of having a standard interface between the agents and the games.  Instead, each game / challenge can adopt the interfaces that best suit it.  Games for the platform can be coded in any programming language, and communication between the game engine and the agents is done by exchanging text messages via console IO.  However, to make a game SFP-ready may mean porting the game to the language of choice and ensuring it has a suitable copy() method.

Codingame is a very popular platform for software developers wishing to practice and achieve recognition for their coding skills while solving game AI challenges.  Codingame also offers a platform for recruiters wishing to hire software developers.  Game AI competitions run on the platform often have thousands of entrants.  On discussion with one of the competition organisers, it is clear that many of the entries are typically just porting or applying standard algorithms, but the leading entries often show significant ingenuity and achieve high standards of game play.  The platform is mentioned here as a possible forum to run wargame AI challenges, and likely generate more entries than for a typical academic competition.

\subsection{Flexibility}

For future research on wargame AI there is a strong need for flexibility.
This includes the ability to change rules, edit scenarios, back-track to a point in a game to consider a new branch. There are also aspects needed to support war games that are not often needed for commercial games.  These include the ability to: moderate manually, set doctrinal restrictions, analyse resulting data for narrative construction and identify and replay critical branch-points.

\subsection{Summary of Platforms}

Of the platforms mentioned above, the ones that are of particular interest for wargame AI are OpenSpiel and GVGAI.  OpenSpiel is well designed and extensible and built for SFP algorithms.
It comes with an extensive set of classic games and a range of recent good performing game AI agents.  However, it does not support structured actions, since each action is coded as an
integer from a (usually small) finite set.  This might sound
like a minor software engineering point, but
is actually a manifestation of a deeper issue
as it affects the operation of most of (perhaps  all) the agents developed for a platform.  For example, many algorithms work by sampling actions
from a relatively small set of possible actions;
if this set is effectively infinite they will
fail unless adapted to cope with this.

GVGAI is of interest due to its use of VGDL, showing a relevant practical example of how a DSL can enable relatively rapid authoring of new games.  Like OpenSpiel though it codes actions as
integers, so for this and other reasons does not
support wargames.

There is much to be gained by developing a new platform built from the ground up for wargames and wargame AI.  This should offer support
for scenario design and analysis, and to
enable easy interaction with a range of AI agents.  This will be discussed in more detail in section~\ref{sect:Recommendations}.

\vfill

\pagebreak

\section{Recommendations/Next Steps} \label{sect:Recommendations}


This section lays out specific recommendations. The main Recommendation 1, and foundation for all the others is to develop a framework suited for wargame AI requirements, and develop all future wargames to this standard. 
Recommendation 2 reviews which challenges are most tractable for AI techniques, and areas to consider for an initial proof of concept.

Recommendations 3a through 3c are outline options to implement AI in wargames based on the analysis in the preceding sections. These are not mutually exclusive, nor are they the only possible options. As Sections~\ref{sect:OtherAlgos} and~\ref{sect:algoSummary} have emphasized, most useful AI solutions are hybrids or iterated developments of others. In this light these recommendations represent three distinct starting points for using AI in wargames that take into account their hierarchical, unit-based structure and have proven their value in recent work.
Recommendation 4 reviews briefly areas of AI research that may be relevant to wargames but which are currently not addressed directly by much of the recent explosion of AI work in Deep RL and elsewhere.

\subsection{Recommendation 1: Develop a new Wargame AI Framework}\label{rec1}

The review of existing platforms in section \ref{sect:Platforms} indicates that the existing game AI platforms do not provide adequate support for wargame AI for a number of reasons. This is not surprising, as none of them were designed with that purpose. 
However, the existing platforms do clearly show the potential 
for a well positioned system to supercharge research in
a particular domain.  A prime example of this is how
the ALE framework (often used via OpenAI Gym) has captured
the imagination of DeepRL researchers especially, leading to
the publication of hundreds of papers.  As a result the
domain of Atari 2600 games is now well understood.

\begin{figure}[h]
\includegraphics[width=1.0\textwidth]{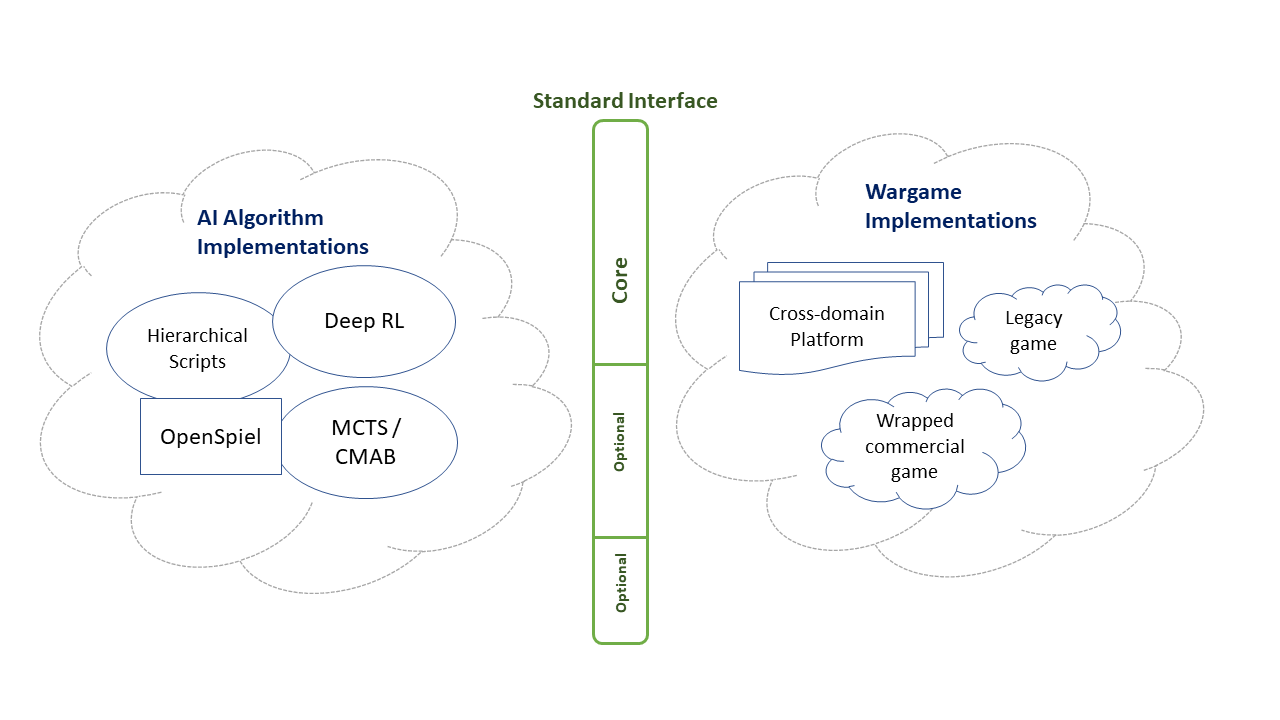}
\caption{Generic proposed architecture with standard interface to cleanly separate AI Algorithm implementations from wargames. Wargames could be based on a new platform, or wrapped legacy/commercial environments. Each wargame would need to implement a minimum Core part of the interface, with support for Optional elements allowing use of increasing number of algorithms and analysis.}
\label{fig:interface}
\end{figure}

Previous attempts have shown that using the latest AI techniques on existing wargame environments can be problematic due to their relatively slow execution time, lack of a clean interface for AI interaction, and lack of a forward model that can be copied or branched easily.
The central recommendation of this report is that a framework is developed to support both the development of AI for wargames, and of wargames that are amenable to AI.
Figure~\ref{fig:interface} shows this high-level architecture, consisting of three conceptual components:
\begin{enumerate}
    \item \textbf{Interface}. This is the core component, and is primarily a design specification rather than software. Like the OpenAI Gym or OpenSpiel interfaces this defines the API methods that a wargame needs to support for AI to be used. This specification can have core methods that must be implemented, and others that are optional. We give some examples of this shortly.
    \item \textbf{Wargames}. Wargame implementations can be new ones, or potentially existing ones that are updated to support the standardised interface.
    \item \textbf{AI Algorithms}. AI development is cleanly separated from the underlying wargames, permitting algorithms to be ported across different environments and new techniques tried. This does not mean that AI development is agnostic to the target wargame, for as this report has catalogued, successful AI techniques are tailored and optimised to the structure of the domain, each of which will have distinctive patterns to its action and observation spaces.
\end{enumerate}


It is possible that existing wargames, either commercial or legacy in-house systems, can be updated to provide an API that supports this new interface, or a software wrapper developed that achieves the same end, although this may require prohibitive work to support all parts of the interface specification, in particular a forward model for SFP algorithms.

One particular recommendation is to develop a software platform for new wargames. This platform can design in from the start key elements useful for the AI interface, such as a fast forward model and hierarchically decomposable observation and action spaces. 
A single platform is also more cost-effective than developing separate wargames in each domain, with some elements, such as units, command structures or enemy visibility common across all types of wargames. 

Using third-party suppliers to wrap/adapt an existing commercial game has some major advantages in terms of leveraging existing functionality and a pool of relevant software engineering expertise. The disadvantages are a potential loss of control over the process and content upgrades. 
The balance between these will obviously depend on the extent to which an existing wargame meets current requirements, and an assessment of the relative depth of skill-sets and experience.

Future defence wargaming environments should be developed to this platform/framework to enable AI techniques to be used uniformly across games. 
While the design framework and interface should be generic, it is advisable to start with an initial domain-specific implementation that meets a current requirement as a proof of concept.
We consider this further in Recommendation 2.

\subsubsection{Interface/Wargame sub-components}
Figure~\ref{fig:interface} suggests that a suitable interface will have core and optional parts, with not all wargames needing to support all of the latter. 
The list below provides a first sketch of interface sub-components, and how these will benefit/impact supporting wargames. This is not an exhaustive list, and the work to detail the requirements beyond this high-level is beyond the scope of this report.

\begin{itemize}
\item \textbf{Core}. A clean design split between the core engine, graphical display, and AI. 
Support for either Deep RL and SFP algorithms requires many game executions, either from start-to-finish or for game-tree planning. Running in `headless' mode with no graphical display or expensive database access is essential to achieve the required speed of execution when there are no humans-in-the-loop. 
The OpenAI and OpenSpiel frameworks provide good model examples, and there is significant overlap but adjustments are needed for wargame-specific features such as multiple units. Some core concepts to consider are:
\begin{itemize}
    \item \texttt{State.step()} to move the time forward one step. In a wargame environment a continuous time argument may make more sense.
    \item \texttt{State.apply(actions)} to specify actions to take. Unlike the OpenAI/OpenSpiel interfaces this should be extended to cover a list of actions for different units.
    \item \texttt{State.registerPolicy(unit, policy)} is an option to provide a policy to be used for one or more agents. An injected policy is then used when the model is rolled forward with \texttt{step()}.
    \item Each \texttt{State} should have, to use the OpenAI terms, an \texttt{action\_space} and \texttt{observation\_space} defined. 
    The \texttt{observation\_space} should be a structured semantic representation and not pixel-based.
    For a wargame support for unit-specific spaces will be needed with a hierarchical structure. 
    It is worth stressing that each wargame will otherwise have different domain-relevant conventions for these spaces, as do the OpenAI Gym environments, or OpenSpiel games. 
\end{itemize}
\item \textbf{Copyable Forward Model}. A \texttt{State.copy()} method to copy the current state to implement a forward model. This is required for support of SFP and other planning algorithms, such as used in AlphaGo. This is put as `optional' here as pure-RL methods do not require it as long as the core game engine is fast (as outlined in Sections~\ref{sect:DeepRL} and~\ref{sect:OtherAlgos}). For existing systems in particular this can be a software engineering challenge to implement with a need for robust deep copies of the game state.
\item \textbf{Observation Filters}. \texttt{State.observationFilter(unit)}-like methods that report only the state visible to a unit, or group of units. This can be enhanced to support different levels of filter to emulate different levels of imperfect information for unit decision-making.
\item \textbf{Belief Injection}. \texttt{State.observe(unit, data)} or \texttt{State.assume(data)} methods to provide API access to the forward model to allow AI algorithms to use the model / engine to evaluate hypothetical plans incorporating estimates about what cannot be currently observed.
\item \textbf{OpenSpiel support}. OpenSpiel provides a number of robust and state of the art Game AI algorithms. An additional interface sub-component could provide conversion support to the OpenSpiel interface to allow these to be used in unit-based wargames.
\item \textbf{Visuals.} \texttt{State.render(unit)} to allow the current state to be visually displayed, perhaps from the perspective of a specific unit or element of the command structure. 
\end{itemize}

If the route to a full wargame software platform is taken, then some key aspects this should include on top of the list above are:
\begin{itemize}
\item Easy generation of features. A feature is a game-specific piece of information likely to be helpful to a decision-making agent. Instead of (or as well as) making the full raw game state available agents, a mechanism to specify these easily for a particular wargame or scenario will speed development and learning.
\item An easy way for users to author new scenarios. This, like the next two points, would be well supported by a Domain-Specific Language (DSL) tailored to wargame requirements. A model is VGDL used by GVGAI discussed in Section~\ref{sect:Platforms}.
\item Extensible: easy to add new unit types
\item Configurable objectives (including multi-objectives)
\item Fully instrumented with configurable logging (this also applies to AI Algorithms so that the data behind each decision is recorded for later analysis and insight mining
\item Record/Replay facility so that a previously played game can be watched later
\item Offer planning at multiple levels of abstraction
\end{itemize}

\subsubsection{Next Steps}
Specifically recommended next steps are:
\begin{enumerate}
    \item Detailed work to define the interface component for wargame AI sketched above. This can, and probably should, be done in parallel with one or both of the following points in an iterative approach.
    \item Identify useful existing wargames that can be adapted without excessive development effort, either in-house or from third-parties. 
    \item Scope possible work on building a more generic wargame platform.
\end{enumerate}

\subsection{Recommendation 2: Tractable Challenges}\label{sect:Challenges}
This section reviews some of the challenges of incorporating the current state of the art AI techniques into wargames, using the framework typology of Section~\ref{sect:wargameTypes}.
In all cases there is an assumption that the relevant sections of the wargame have been engineered in line with the recommendation in Section~\ref{rec1}. 

Underlying conclusions are that using AI to decide on communication with other agents or humans is at the very challenging end.
A general rule in all cases is that the smaller the scenario, and less of a strategic master-plan is required, the more suitable AI will be. For example a sub-hunt scenario that involves 2-3 destroyers is tractable, as is a company-level scenario to secure and hold an objective. A scenario that models an amphibious assault on an extensive archipelago with sea and air support is less suitable.
This is particularly true for an initial proof of concept. 

From the review below, some potential scenario-agnostic goals for a proof of concept are:
\begin{enumerate}
    \item Unit `micro-control' as assistance for human players in a computer-moderated wargame. This avoids the need for strategic level decision-making, which is the responsibility of the human player. An analogy is the in-game AI in Flashpoint Campaigns that controls the detailed moves of the player's units. The complexity can be reduced if each unit is autonomous, making decisions based primarily on local information as in DecPOMDP models.
    \item AI full control of all units in a small wargame, with a single clearly defined objective. Either for use as an opponent for human players, or for analyst insight with scenario results with both sides under AI control. Adding realism to the AI through full imitation of human-like behaviours would add considerable complexity. More realistically different reward functions could be tailored to obtain behavioural variations.
    \item AI control of one or both sides in a Concept Development wargame. In this context the open-ended goal of seeking a diverse range of strategies can be supported by use of multi-objective approaches. The ability with AI to run hundreds or thousands of games with the same, or slightly different setup is useful for statistical analysis of the results, as well as permitting a full exploration of the strategy space.
    \item As the previous point, but in a Procurement setting. This also benefits from the ability to run many games to feed into statistical analysis, but the requirement to have control experiments with doctrinal realism of behaviour adds challenge if this cannot be easily codified.
\end{enumerate}

\subsubsection{Planned Force Testing}
Significant incorporation of AI into large-scale Planned Force tests involving large numbers of communicating groups and sub-teams is not a realistic goal in the immediate future. This requires breakthroughs in areas of modelling restricted information flow and co-ordination between multiple agents not addressed seriously be recent research (see Sections~\ref{sect:InfoFlow} and~\ref{sect:AdvOpponent}). The scope of including diplomatic and/or external logistic aspects is also not well studied.

However, there may be some opportunities to incorporate AI in aspects of the moderation of parts of a large-scale wargame, which are covered in the following sections.

\subsubsection{Plan Variation Testing}\label{sect:PlanVarTest}
Given a specific scenario to be repeated a number of times there are a few ways that AI techniques could be utilised, of varying challenge levels. Two aspects are specifically relevant here: the level of control the AI has, and the constraints under which it must operate.
\begin{enumerate}
    \item{\textbf{AI control}
\begin{itemize}
\item Interactive control. The AI is expected to emulate the dynamic behaviour of a human player by responding to changing objectives and orders as the scenario unfolds. Two-way communication, responding to orders or providing an interpretative summary of the current situation is not feasible. One-way communication with precise changes to strategic targets is more feasible, as long as these can be encoded clearly in a utility or state evaluation function, but still very challenging. Any adaptive requirement here would be much easier to support with SFP or other planning methods, or else known changes would need to be included in pre-training for purer RL approaches. 
\item Full control. The AI has full control of one side in the game, but with a fixed objective at the start of the scenario that does not change (although clearly the AI must still respond to the actions of the opposing side). This is less challenging than interactive control, and is less restrictive about applied methods with pure RL now competitive. It is also most similar to the Dota2 and Starcraft II game set-ups, which set a rough upper bound on the challenge and cost level.
\item Unit micro-management. A human player is responsible for strategic decisions, while the AI is responsible for moving units on a hex-by-hex basis to implement these. This would be most similar, for example, to the role of the AI in Flashpoint Campaigns, which decides for each unit exactly where to move, and when to attack visible enemies. The benefit sought here is to make best use of the human player's time, and avoid it being spent on low-value activities. The challenge in this case is reduced significantly by the lack of need to consider strategic moves.
\end{itemize}
}
\item{\textbf{AI constraint}
\begin{itemize}
\item Unit independence. Each unit (or group of units) is controlled independently, so only acts based on its local knowledge along with some initially specified goal. This contrasts with a default approach in which a single AI controls all units, taking into account what each of them can see. The local, independent approach is less computationally challenging, as it avoids some of the problems of large action and state spaces. It would require some additional features in the API discussed in Section~\ref{rec1} to support it. The default approach of all units being controlled by one AI is simpler architecturally, and theoretically more optimal, but is not computationally tractable for large numbers of units. 
\item Fog of War. Fully intelligent understanding of imperfect information is challenging, although the support of it is not, provided this is factored up-front into the API design. By `intelligent' we mean taking into account the value of information-gathering moves, and bluffing/second-guessing the opponent (Section~\ref{sect:Observability}). However, a good simulation of an intelligent understanding can be obtained by crafting the reward function to include a bonus for visible terrain, and scouting activities.
\item Military doctrine. Where an AI needs to act in line with specific tactical doctrine to provide useful results, the challenge is clearly defining what the military doctrine is. If this can be defined in game mechanic terms, then this can be implemented by filtering non-compliant actions from the action space. More strategic elements can be factored in by engineering the reward function or victory points awarded. Learning a doctrine implicitly from observations of human actions is much more difficult.
\item Specific Plan. More challenging is where the AI needs to broadly follow a plan generated as an outcome from a previous physical wargame, minimising variations from this. This can potentially be supported by incorporating the target plan into the reward function, although with less weight than the actual victory conditions. This will incentivise the AI to keep as closely as possible to the plan, but not at the expense of winning. Deviations from the target plan can then be analysed for common breakpoints.
\end{itemize}
}

\end{enumerate}
\subsubsection{Concept/Force Development}\label{sect:ConceptDev}
These wargames are particularly suitable for using AI due to their explicitly exploratory goals. These means less work is required to impose constraints based on tactical doctrine or \textit{a priori} assumptions on the best/correct way to act in a situation, and algorithms such as MAP-Elites and multi-objective training are especially useful here to encourage a diverse array of solutions (see Section~\ref{sect:Exploration}).
The issues instead revolve around interpretability of the results and insight extraction. 
\begin{itemize}
    \item \textbf{Human Observation}. Analyst observation of games using the final learned policies can be a prime source of insight.
    \item \textbf{Statistical Analysis}. This requires flexible and in-depth instrumentation as discussed in Section~\ref{rec1}. This focuses on patterns in the course of game, and does not provide direct information on why those patterns occur. This would still require analyst insight.
    \item \textbf{Plan interrogation}. Instrumentation can include all the actions an agent considers at each step, and the relative weights applied to each; for example the visit counts in MCTS or the Q-value of the action in Reinforcement Learning. In the case of SFP methods this query can be extended deeper into future simulated at that decision point to understand what the agent believed would happen to some time horizon. This can highlight critical decision points where different actions had very similar values, but led to very different future results.
    \item \textbf{Feature Perturbation}. Plan interrogation does not provide direct evidence as to why an agent took a particular decision. Feature Perturbation is especially helpful in Deep RL methods in which the neural network that makes a decision is very difficult to interpret directly. This modifies different inputs to the decision and highlights those that were instrumental in making the decision~\cite{Gupta_Puri_Verma_Kayastha_Deshmukh_Krishnamurthy_Singh_2020}. See Section~\ref{sect:ExplainableAI} for further methods.
\end{itemize}

\subsubsection{Procurement Testing}
The smaller scale of these wargames makes these amenable to AI techniques, and many of the points made in Sections~\ref{sect:ConceptDev} and~\ref{sect:PlanVarTest} apply here. Statistical analysis from multiple iterations (assuming AI versus AI games) comparing the outcomes when equipment/doctrine is varied between a small number of specific options would provide further benefits beyond what is currently achievable due to resource constraints with human-in-the-loop games that limit the number of iterations.
The main challenge in this domain is applying doctrinal and other constraints to AI behaviour.

\subsection{Recommendation 3a: A Scalable Deep Learning Approach for Wargaming through Learned Unit Embeddings}
This section presents one particular deep learning approach that we believe is promising for wargaming. Wargames come with their own particular requirements and challenges compared to other games that deep learning approaches have been applied to so far. Here we propose one particular deep learning approach that combines the most relevant aspects of the AlphaStar and OpenFive architecture, while paying particular attention to the computational and scaling demands of the approach.  

\begin{figure}[h]
\includegraphics[width=1.0\textwidth]{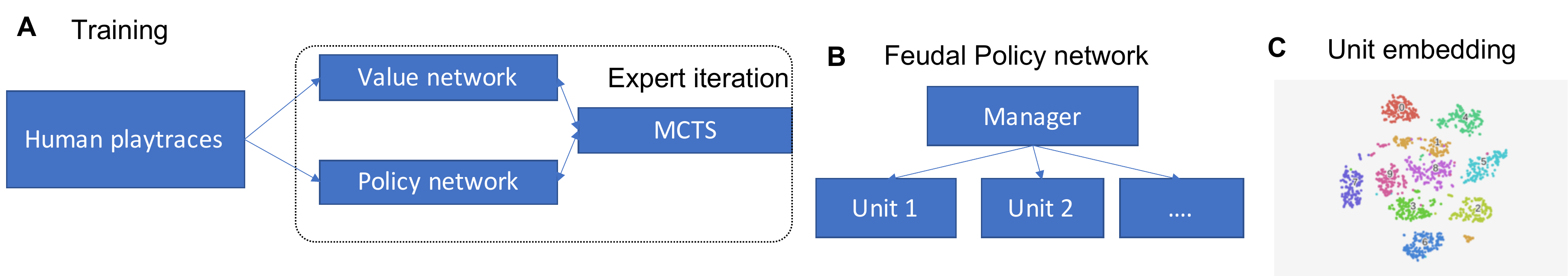}
\caption{Training (A). A policy network that predicts a distribution of valid moves and a value network that predicts the expected game outcome are first trained based on human examples. Policies are further fine-tuned through expert iteration. The policy network (B) is a feudal network in which a manager controls lower level units. The manager and unit network takes as input processed unit information such as distances, enemy types, etc. instead of working from raw pixels. (C) In order to support new unit types without having to retrain the whole  system, unit types embeddings are based on unit abilities. This way the feudal network should be able to generalise to new units, based on similar units it already learned to control. }
\label{fig:approach1}
\end{figure}

An overview of the proposed approach is shown in Figure~\ref{fig:approach1}. Similarly to AlphaStar or AlphaGo, we first train a value and policy network based on  existing human playtraces in a supervised way. First training on existing playtraces, instead of starting from a \emph{tabula rasa}, will significantly decrease the computational costs needed for learning a high-performing policy. In the second step, and given that a fast forward model of the game is available, the policy can be further improved through an expert iteration algorithm (see Figure~\ref{fig:ExpertIteration}). 

In terms of employed policy network, we believe the most scalable version with the least computational demands would be a Feudal network \cite{ahilan2019feudal} approach (Figure~\ref{fig:approach1}b).  Similarly to a system such as AlphaGo, one manager network controls each of the different units. One main disadvantage of the AlphaStar system for  wargaming is that it would require retrained from scratch every time a new unit or weapon type is introduced. Therefore we propose a new Feudal policy manager approach that does not directly work with specific unit types, but with \emph{unit embeddings} instead. Embeddings are most 
often used in a subfield of the deep learning dealing with natural language processing \cite{levy2014dependency}. The idea is that words that have a similar meaning should be represented by a learned representation of vectors that are similar to each other. We imagine a similar embedding can be learned for different unit and weapon types. This way, the Feudal policy network would not have to be retrained once a new unit is introduced and the network should be able to interpolate its policy to a new unit, whose embedding should be close to already existing unit types.

Other approaches such as (1) automatically learning a forward model for faster planning in latent space (Section~\ref{sec:learning_fm}), or (2) adding functionality for continual learning (Section~\ref{sec:continual_learning}) could be added to this system in the future. To train such as system, we believe the Fiber framework (Section~\ref{sec:deep_rl_frameworks}) introduced by Uber could be a good fit and would require minimal engineering efforts.

Given a fast enough simulator and the fact that we should be able to kickstart training with human replays, we speculate that even a modest setup of $\sim$100 CPUs and  $\sim$5 GPUs might allow the system to be trained in around 12 hours.



\subsection{Recommendation 3b: Graph Neural Networks for Unit Information Integration}\label{sect:GNN}
\begin{figure}[h]
\centering
\includegraphics[width=0.5\textwidth]{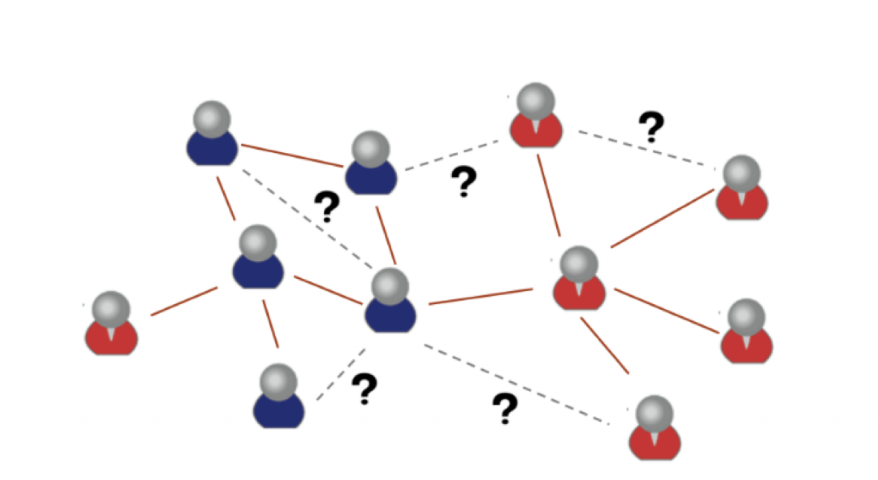}
\caption{Example of a Graph Neural Network modelling a social network \cite{zhou2018graph}. A similar structure could model the chain of command and communication in wargames.}
\label{fig:gnn}
\end{figure}

In addition to using mostly off-the-shelf components as in described in the previous section, a less explored but potentially very relevant approach for wargaming are graph neural networks. Graph neural networks (GNN) are becoming more and more popular in deep learning \cite{zhou2018graph}, and have shown promising results in domains such as physics system, learning molecular fingerprints, modelling disease, or predicting protein interactions. The basic idea behind graph neural network (Figure~\ref{fig:gnn}), is to model the dependencies of graphs via learned message passing between the nodes of the graph. For example, a Sudoko puzzle can be modelled as a graph \cite{palm2018recurrent}, in which each of the 81 cells in the 9$\times$9 Sudoko grid is a node in the graph. The same neural network is used for all nodes in the graph to learn to integrate information from incoming nodes and to iteratively come up with the Sudoko solution.

In the case of wargaming, nodes in the GNN could represent units, while the connections between them reflect communication channels. A GNN could be trained to integrate information in a decentralised way and learn by itself to resolve communication issues and misalignment (e.g.\ seeing the same unit twice,  noisy communication channels) into a coherent whole. Additionally, a particularly relevant and recently proposed GNN architecture is called recurrent independent mechanism (RIM) \cite{goyal2019recurrent}. In this system, units only sporadically communicate and keep a  level of autonomy. These RIMs might be particularly promising to model situations in which a lower level wargaming unit lost contact to a higher level orders. The unit would have to learn to work both when higher levels orders are available, but also compensate and follow their own autonomy when they are not.

\subsection{Recommendation 3c: High Performance Statistical Forward Planning Agents} \label{HighPerfSFP}


As an alternative/parallel approach to the Deep RL-focus of the previous two recommendations, an SFP approach could productively be built on recent results in RTS and multi-unit game environments.
This approach would leverage domain knowledge in the form of low-level action scripts to provide Action Abstraction. Domain knowledge can also be incorporated in a state evaluation function used to determine the value of an action choice after rolling out the forward model.

Specific variants here are:
\begin{enumerate}
    \item A Combinatorial Multi-Armed Bandit approach, such as NaiveMCTS~\cite{Ontanon_2017} or direct MCTS~\cite{Churchill_Buro_2015}. The expert-scripts provide the set of actions for each unit. This is suitable for a relatively small number of units.
    \item A two-stage approach that uses a neural network trained by RL to make the initial decision for each unit, followed by low-level modification of this plan using MCTS or CMAB as in~\cite{Barriga_Stanescu_Buro_2017} to take account of unit (and enemy) interactions. This makes more effective use of the computational budget to scale up to larger scenarios.
    \item Online Evolutionary Planning (OEP/RHEA) as an evolutionary option for multiple units~\cite{Justesen_Mahlmann_Risi_Togelius_2018}.
    \item Stratified Strategy Selection~\cite{Lelis_2017}. All units are partitioned into a set of types, for example based on position, unit type, damage and so on. Each type is then given a single consistent order (i.e. script), and the algorithm searches in the space of type partitions and script combinations. The scales well to large numbers of units. 
\end{enumerate}

In all cases the Action Abstraction scripts and/or the state evaluation functions can be learned through either Evolutionary methods (see~\cite{Marino_Moraes_Toledo_Lelis_2019, Neufeld_Mostaghim_Perez-Liebana_2019}), or Expert Iteration based on RL from expert games as described in Section~\ref{sect:OtherAlgos}.

\subsection{Recommendation 4: Open areas of research}
As this report has shown, much of the intensive research effort over the past few years into Deep RL and Game AI has generated techniques that are selectively very relevant to wargames.
Below we summarise a number of areas relevant to wargames that have not benefited from this surge of recent work, and which are relatively under-developed.
\begin{itemize}
    \item Restricted Information Flow between agents, and between low-level units and high-level commanders. 
    \item Diplomacy and Negotiation in games with more than two players.
    \item Exploring Graph Neural Networks for modelling unit information integration
    \item Combining deep RL with learned unit embeddings as a scalable approach for wargaming AI
    \item Integration of human-factor research on realistic behavioural patterns of military commanders into AI constraints
\end{itemize}

\vfill


\section{Conclusions}

Great strides have been made in Game AI over the last 5 years in particular, though the headline achievements using DeepRL methods have often been achieved at the cost of great effort, expertise and expense.

Wargames are in many cases not fundamentally harder than playing StarCraft to a high standard, but they are different to commercial games and come in a wide variety.  In most cases it will not be possible to put vast amounts of effort into developing strong AI for any particular wargame, but there remain many interesting and open questions regarding the level of competence that could be achieved using existing recent methods, and the amount of investment needed to achieve satisfactory performance.

We make a number of recommendations that offer viable ways forward and we firmly believe that these should be explored in detail and taken forward as appropriate.  There has never been a better time to invest in this area, and the potential benefits are significant.  They include better staff training, better decision making, and an improved understanding of the nuances of problem domains that comes as an inevitable and desirable spinoff from building better models and gaming with them.

\vfill

\pagebreak

\small

\bibliographystyle{abbrv}
\bibliography{bibliography}

\end{document}